\documentclass[10pt,journal,compsoc]{IEEEtran}
\usepackage{graphicx}
\usepackage{algorithmicx}
\usepackage{algorithm}
\usepackage{algpseudocode}
\usepackage{bm}
\usepackage{multirow}
\usepackage{amsmath,amssymb,amsfonts}
\usepackage{amsthm}
\newtheorem{theorem}{\bf \emph{Theorem}}

\newtheorem{Definition}{\bf \emph{Definition}}
\usepackage{xcolor}

\usepackage{makecell}



\ifCLASSOPTIONcompsoc
  \usepackage[nocompress]{cite}
\else
  \usepackage{cite}
\fi

\begin{document}

\title{How to optimize K-means?}

\author{Qi~Li
\IEEEcompsocitemizethanks{\IEEEcompsocthanksitem Q. Li is with School of Information and Technology, Beijing Forestry university, Beijing, 100083, China\protect\\
E-mail: liqi2024@bjfu.edu.cn}
}

\IEEEtitleabstractindextext{%
\begin{abstract}
Center-based clustering algorithms (\emph{e.g.}, K-means) are popular for clustering tasks, but they usually struggle to achieve high accuracy on complex datasets. We believe the main reason is that traditional center-based clustering algorithms identify only one clustering center in each cluster. Once the distribution of the dataset is complex, a single clustering center cannot strongly represent distant objects within the cluster. How to optimize the existing center-based clustering algorithms will be valuable research. In this paper, we propose a general optimization method called ECAC, and it can optimize different center-based clustering algorithms. ECAC is independent of the clustering principle and is embedded as a component between the center process and the category assignment process of center-based clustering algorithms. Specifically, ECAC identifies several extended-centers for each clustering center. The extended-centers will act as relays to expand the representative capability of the clustering center in the complex cluster, thus improving the accuracy of center-based clustering algorithms. We conducted numerous experiments to verify the robustness and effectiveness of ECAC. ECAC is robust to diverse datasets and diverse clustering centers. After ECAC optimization, the accuracy (NMI as well as RI) of center-based clustering algorithms improves by an average of 33.4\% and 64.1\%, respectively, and even K-means accurately identifies complex-shaped clusters.
\end{abstract}

\begin{IEEEkeywords}
clustering, optimization, clustering centers, extended-centers
\end{IEEEkeywords}}

\maketitle

\IEEEdisplaynontitleabstractindextext

\IEEEpeerreviewmaketitle

\IEEEraisesectionheading{\section{Introduction}}

Clustering is an important data analysis technique that divides objects into different categories solely based on similarity. Clustering has been widely used in many fields, such as pattern recognition, data compression, image segmentation, time series analysis, information retrieval, spatial data analysis, biomedical research, and so on \cite{wang2016automatic}.

So far, thousands of clustering algorithms have been proposed successively, among which center-based clustering algorithms are the most common \cite{jain2010data}. The most famous clustering algorithm, K-means \cite{macqueen1967classification}, and DPC \cite{rodriguez2014clustering}, which has been widely concerned in recent years \cite{wang2021extreme, lotfi2020density, seyedi2019dynamic, pizzagalli2019trainable, du2018robust, liu2018shared, zhou2018robust, ding2017entropy, mehmood2016clustering, du2016study, xie2016robust, yaohui2017adaptive, lotfi2016improved}, are both center-based clustering algorithms. Traditional center-based clustering algorithms consider that there is a unique object in each cluster that can represent other objects within the cluster, and the object is called the clustering center. In general, center-based clustering algorithms consist of two separable processes, the center process and the category assignment process. The center process identifies the most representative object in the dataset as the clustering center. For example, the center process of K-means identifies the local center object as the clustering center, and the center process of DPC identifies the object with the local density peak as the clustering center. The category assignment process assigns other objects to different clustering centers to form clusters (\emph{i.e.}, categories). Although center-based clustering algorithms are frequently used in real-world scenarios, they generally perform poorly on the datasets with complex distributions. How to improve the accuracy of traditional center-based clustering algorithms on complex datasets will be valuable research.

We believe that the main reason for the poor performance of traditional center-based clustering algorithms is that the representative capability (See Section \ref{sec:Overview-ECAC} for details) of clustering centers is insufficient. Once the distribution of the dataset is complex, it is difficult for a single clustering center to represent distant objects within the cluster. Therefore, in this paper, we especially propose an optimization method called ECAC (Identifying \textbf{E}xtended-\textbf{C}enter to improve the \textbf{A}ccuracy of \textbf{C}enter-based clustering algorithms) for traditional center-based clustering algorithms to improve the representative capability of clustering centers. ECAC is embedded between the center process and the category assignment process, so it does not require adaptive modifications to center-based clustering algorithms. As long as center-based clustering algorithms can be divided into the center process and the category assignment process, ECAC can optimize them. In other words, ECAC is a general optimization method. Specifically, after the center process identifies all clustering centers, ECAC extracts some high-density objects as extended-centers until extended-centers are spread across the entire dataset. The extended-centers derived from the same clustering center are collected as an extended-set. Next, ECAC inputs clustering centers and extended-centers into the category assignment process, resulting in numerous initial-clusters. Finally, according to the extended-set, initial-clusters are merged to obtain clustering result. Since the extended-centers act as relays to strengthen the representative capability of a single clustering center within each cluster, the accuracy of the center-based clustering algorithm optimized by ECAC can be effectively improved. We conducted a series of ablation experiments to validate and discuss the necessity of the components of ECAC. In addition, we also conducted extensive experiments to test its robustness and effectiveness. Experimental results show that ECAC can effectively identify extended-centers in Gaussian datasets, overlapping datasets, shaped datasets, and density-unbalanced datasets. Each extended-set corresponds to only one cluster. After ECAC optimization, the accuracy of the three common center-based clustering algorithms, K-means, DPC, and Extreme clustering, has been significantly improved, and even K-means can accurately identify non-spherical clusters.

We summarize the main contributions of this work as follows: 
\begin{enumerate}
\item We are the first to focus on how to improve the accuracy of traditional center-based clustering algorithms without modifying the clustering principle.

\item We design a general technical route, using numerous extended-centers derived from clustering centers as relays to improve the representative capability of a single clustering center within each cluster, thereby improving clustering accuracy.

\item We propose the ECAC method based on the designed technical route, theoretically demonstrate its feasibility, and discuss its effectiveness in detail.

\item ECAC can be used to optimize different center-based clustering algorithms.

\item ECAC performs well, greatly improving the clustering accuracy (NMI as well as RI) of center-based clustering algorithms (average improvement ratio of  33.4\% and 64.1\%), and even enabling K-means to accurately identify non-spherical clusters.
\end{enumerate}

The remainder of the paper is organized as follows. The next section is about related works. Section \ref{sec:ECAC} introduces the theory of ECAC. Section \ref{sec:Experiments} verifies the robustness and effectiveness of ECAC. The conclusion is reported in Section \ref{sec:Conclusion}.

\section{Related Works}
\label{sec:Related-Works}
Traditional center-based clustering algorithms can be roughly classified into two classes. The first-class algorithms identify clustering centers based on distance, and the second-class algorithms identify clustering centers based on density.

Among the first-class algorithms, K-means \cite{macqueen1967classification} is the most representative and it is also the most well-known and frequently used clustering algorithm. K-means first selects $k$ objects as the initial clustering centers and assigns the other objects to the nearest initial clustering center. Then, in each cluster, it identifies the object closest to other objects as a new clustering center and iterates this process until the clustering center no longer changes. K-means is very simple and has low computational complexity. Therefore, although it has been proposed for more than 60 years, it still receives a lot of attention \cite{jain2010data} and its improved versions are still emerging today \cite{huang2021robust, yu2021dynamic, xia2020fast, moshkovitz2020explainable, alswaitti2018optimized, khanmohammadi2017improved, kumar2017efficient, capo2017efficient, guo2017deep}. WK-means \cite{huang2005automated} considers the features of objects as inequalities. It automatically learns feature weights during the iteration of clustering centers. OGC \cite{alswaitti2018optimized} redesigns the clustering center iteration rule, and it forces non-clustering centers to pull the clustering centers to move to obtain the best clustering centers. KHM-OKM \cite{khanmohammadi2017improved} introduces $k$-harmonic means to solve the sensitivity of the initial clustering centers. RDBI \cite{kumar2017efficient} retrieves the distance between objects based on the KD-tree, thus speeding up the iteration of new clustering centers. RPKM \cite{capo2017efficient} divides the dataset into numerous thin partitions, each characterized by centers and weights. To minimize the number of computed distances, it performs weighted K-means on each thin partition. K-means and deep learning are combined in RDKM \cite{huang2021robust}, in which K-means is performed hierarchically in a deep structure. After the multi-layer transformation, the objects from the same cluster are close to one another. When K-means determines the clustering centers, DCEC \cite{guo2017deep} trains a convolutional autoencoder network until each non-clustering center can be explicitly assigned to a clustering center. SGC \cite{long2006new} forces similar objects to approach and be merged. Finally, the remaining objects can be used as the initial clustering centers of K-means. 

Among the second-class algorithms, DPC \cite{rodriguez2014clustering} is the most representative. It is another milestone clustering algorithm after K-means and DBSCAN \cite{ester1996density}. During this decade, hundreds of its improved versions have been proposed one after another \cite{wang2021extreme, lotfi2020density, seyedi2019dynamic, pizzagalli2019trainable, du2018robust, liu2018shared, zhou2018robust, ding2017entropy, mehmood2016clustering, du2016study, xie2016robust, yaohui2017adaptive, lotfi2016improved}. DPC designs two metrics, local density and the minimum distance to objects with higher density. Based on the two metrics, it generates a decision graph. From the decision graph, the object with the highest density in each cluster can be visually identified as the clustering center. Finally, the non-clustering centers are assigned to the clustering centers along the density gradient direction. DPC-KNN \cite{du2016study} introduces KNN (K-Nearest Neighbors) and PCA (Principal Component Analysis) to DPC to reduce the complexity of density calculation. On the other hand, SNN-DPC \cite{liu2018shared} replaces the cut-off distance with the share neighbor information to improve the robustness of density calculation. Extreme clustering \cite{wang2021extreme} identifies density extremes as clustering centers, allowing it to cope with density-unbalanced datasets. The clustering accuracy of Extreme clustering is substantially higher than that of DPC. FN-DPC \cite{du2018robust} utilizes fuzzy neighbor metric to identify high-density objects. It can accurately identify only one clustering center in each shaped cluster, so solves the defect that DPC is not friendly to shaped clusters. FKNN-DPC \cite{zhou2018robust} alters the category assignment strategy of DPC. It assigns non-clustering centers through the breadth-first search of nearest neighbors. DPC-SP \cite{pizzagalli2019trainable} replaces local association rules with global optimization. It assigns non-clustering centers to clustering centers along the shortest path. 

\begin{figure*}
  \centering
  \includegraphics[width=5in,height=2in]{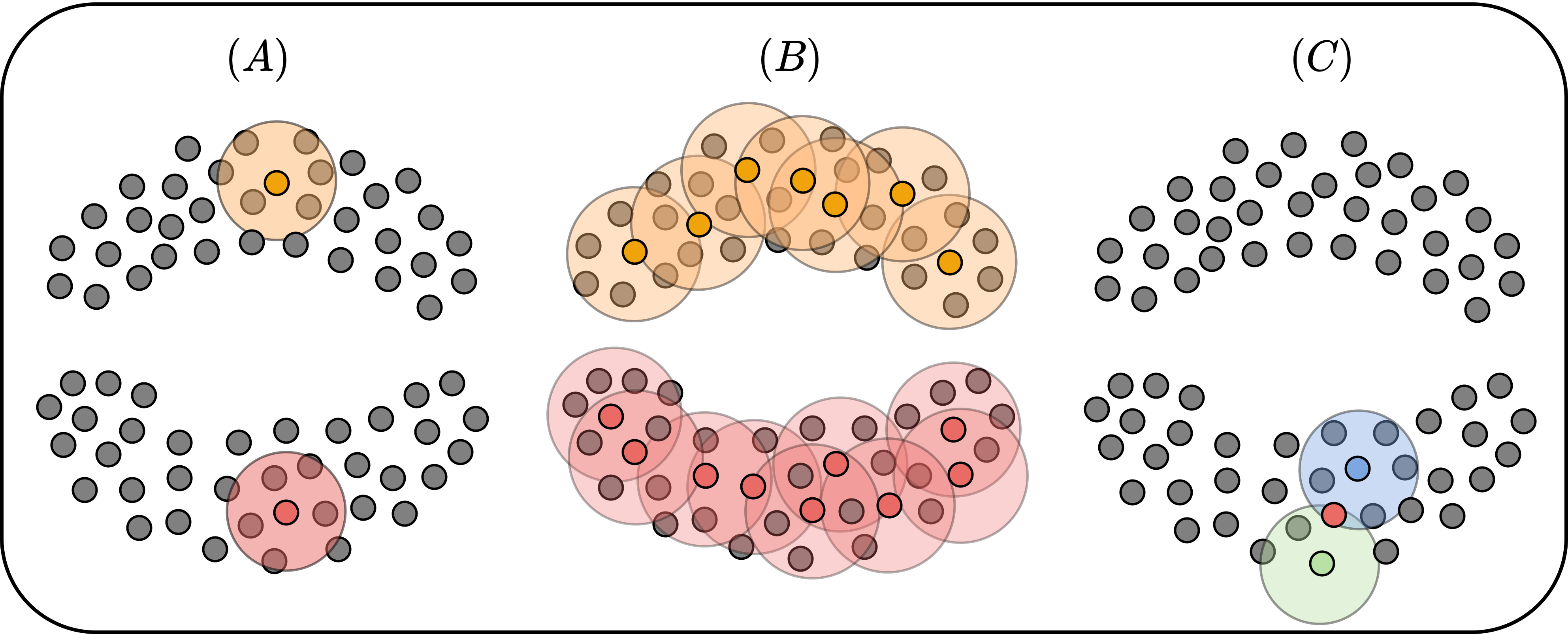}
  \caption{\textbf{An example of ECAC:} A) A single clustering center has insufficient representative capability to cover the entire cluster; B) Each clustering center shares the representative degree of its derived extended-centers, making its representative capability cover the entire cluster; C) choosing high-density objects as extended-centers will increase more representative capability.}
  \label{fig:example}
\end{figure*}

\section{The proposed method: ECAC}
\label{sec:ECAC}
\subsection{Overview of ECAC}
\label{sec:Overview-ECAC}
Before introducing ECAC, we first define two concepts, representative degree and representative capability:
\begin{Definition}
\label{Definition1}
\textbf{(Representative degree):} The similarity between object $o_i$ and object $o_j$ is called the representative degree of object $o_i$ to object $o_j$.
\end{Definition}

\begin{Definition}
\label{Definition2}
\textbf{(Representative capability):} The number of objects, which $o_i$ has high representative degree to, is called the representative capability of object $o_i$.
\end{Definition}

It is well known that similarity is inversely proportional to the distance. Therefore, the representative degree is also inversely proportional to the distance. The farther the distance between object $o_i$ and object $o_j$ is, the smaller the representative degree of object $o_i$ to object $o_j$ is. As shown in Figure \ref{fig:example}(A), there are two clusters, in which the yellow object and the red object are the clustering centers of the upper and lower clusters, respectively. Apparently, the two clustering centers have higher representative degree to the objects within the shaded area than the objects outside the shaded area. If the shaded area is used as the boundary for distinguishing high and low representative degree, then the number of the objects within the shaded area is the representative capability of the two clustering centers. Obviously, their representative capability is not enough to cover the entire dataset. Therefore, we can conclude that the accuracy of center-based clustering algorithms can be improved if the representative capability of the clustering center is improved. 

In this paper, we propose an optimization method called ECAC (Identifying \textbf{E}xtended-\textbf{C}enter to improve the \textbf{A}ccuracy of \textbf{C}enter-based clustering algorithms) for traditional center-based clustering algorithms. We try to select some ordinary objects as the extended-centers of the clustering centers. These extended-centers will act as relays to expand the representative capability of clustering centers. In this way, ECAC can greatly improve the accuracy of center-based clustering algorithms without modifying the clustering principle. We divide the optimization into 3 steps (the first step is not related to ECAC):
\begin{enumerate}
\item[1.] {\textbf{Identifying clustering centers:} The center process of center-based clustering algorithms identifies clustering centers from the dataset.}
\item[2.] {\textbf{Identifying extended-centers:} ECAC identifies extended-centers from ordinary objects. The extended-centers are grouped into different extended-sets according to their derivation sources (\emph{i.e.}, clustering centers).}
\item[3.] {\textbf{Identifying clusters:} ECAC inputs both clustering centers and extended-centers into the category assignment process of center-based clustering algorithms to obtain the initial-clusters. Finally, ECAC merges these initial-clusters according to extended-sets to obtain the optimized clustering results.}
\end{enumerate}

\subsection{Introduction of notations}
$O$ represents the dataset to be analyzed, and its size is $N$. The $i$-th object in $O$ is denoted as $o_i$. $c_i$ represents the $i$-th clustering center identified by the center process. $E$ is a set storing clustering centers and extended-centers, and $E_i$ represents the extended-set where $c_i$ is located. The Euclidean distance between object $o_i$ and object $o_j$ is denoted as $\|o_i-o_j\|_2$. $\delta(o_i)$ represents the neighborhood of $o_i$ with the radius $\delta$, $\delta\left(o_i\right)=\{o_j\in O|\|o_i-o_j\|_2<\delta\}$. The center process of center-based clustering algorithms is denoted as $fun\_center()$, and the category assignment process is denoted as $fun\_label()$.

\subsection{Identifying clustering centers}
For the dataset $O$, let us assume that a center-based clustering algorithm identifies $k$ clustering centers during its center process,
\begin{equation}
\left\{c_1,c_2,\cdots,c_k\right\}=fun\_center(O).
\label{eq:clustering centers}
\end{equation}
Next, we will derive numerous extended-centers based on these $k$ clustering centers to form $k$ extended-sets, $E_1, E_2, \cdots, E_k$.

\subsection{Identifying extended-centers}
\label{sec:extended-centers}
In the initial state, $E_i=\left\{c_i\right\}$. We try to identify the high-density objects closest to extended-sets as extended-centers (\textbf{We will explain why we select them as extended-centers in Section \ref{sec:Feasibility}}), so the distance between ordinary object $o_i$ and extended-set $E_j$ is defined as
\begin{equation}
dis(o_i,E_j)=\frac{\mathop{\min}\limits_{x\in E_j}{\|o_i-x\|_2}}{\rho(o_i)}
\label{eq:dis}
\end{equation}
, in which $\rho\left(o_i\right)$ is the density of object $o_i$, $\rho\left(o_i\right)=\sum_{j=1}^{N}{\varphi\left(o_j\middle|\delta(o_i)\right)}$. If $o_j\in\delta(o_i)$, $\varphi\left(o_j\middle|\delta(o_i)\right)=1$; otherwise, $\varphi\left(o_j\middle|\delta(o_i)\right)=0$. $\delta$ is the only input parameter of ECAC, and it should be set to a smaller value. \textbf{We will demonstrate the advantage of the formula (\ref{eq:dis}) in Section \ref{sec:Feasibility} and experimentally verify these advantages in Section \ref{sec:density}.} Next, we perform the following steps:
\begin{itemize}
\item{\textbf{Step 1:} $E=\left\{c_1,c_2,\cdots,c_k\right\}$, $\Delta=\mathop{\bigcup}\limits_{x\in E}{\delta(x)}$. Since the value of $\delta$ is small, the object $x$ has high representative degree to the objects within $\delta(x)$. Therefore, the size of $\Delta$ can be regarded as the representative capability of the clustering centers in $E$.}

\item{\textbf{Step 2:} Find the pair with the smallest distance between objects and extended-sets, $o_\Theta,E_\dag=\mathop{\arg\min}\limits_{o_i\in O-E, j\in\{1,2,\cdots,k\}}{dis(o_i,E_j)}$.}

\item{\textbf{Step 3:} Treat $o_\Theta$ as an extended-center of clustering center $c_\dag$, and expand the extended-set $E_\dag$, $E_\dag=E_\dag\cup\left\{o_\Theta\right\}$.}

\item{\textbf{Step 4:} $E=E\cup\left\{o_\Theta\right\}$, $\Delta=\Delta\cup\delta(o_\Theta)$.}
\end{itemize}
Repeat Steps 2 to 4 until $\Delta=O$ (\emph{i.e.}, the representative capability of clustering centers and extended-centers covers the entire dataset). It should be noted that, to reduce running time, we propose to search for new extended-centers only within the neighborhood of identified extended-centers with radius $2\delta$ (\emph{i.e.}, Step 2 is modified to $o_\Theta,E_\dag=\mathop{\arg\min}\limits_{o_i\in (\mathop{\bigcup}\limits_{x\in E}{2\delta(x)})-E, j\in\{1,2,\cdots,k\}}{dis(o_i,E_j)}$). \textbf{We will experimentally verify the feasibility of this proposal in Section \ref{sec:local}}.

\subsection{Identifying clusters}
Let us assume that there are $s$ objects in the final $E$, $E=\left\{e_1,e_2,\cdots,e_s\right\}$ $(k\ll s, \left\{c_1,c_2,\cdots,c_k\right\}\subset E)$. ECAC inputs these $s$ objects into the category assignment process of center-based clustering algorithms to obtain $s$ initial-clusters,
\begin{equation}
clu(e_1),\cdots,clu(e_i),\cdots,clu(e_s)=fun\_label(O,E)
\label{eq:initial clusters}
\end{equation}
, in which $clu(e_i)$ is the $i$-th initial-cluster where $e_i$ ($e_i$ is a clustering center or an extended-center) is located. Finally, ECAC merges these $s$ initial-clusters into $k$ final clusters according to extended-sets. The $i$-th final cluster, in which the clustering center $c_i$ is located, is as follows
\begin{equation}
cluster(c_i)=\mathop{\bigcup}\limits_{x\in E_i}{clu(x)}  
\label{eq:final clusters}
\end{equation}
. $\{cluster(c_1),cluster(c_2),\cdots,cluster(c_k)\}$ is the clustering result of the center-based clustering algorithm optimized by ECAC.

\subsection{Feasibility analysis}
\label{sec:Feasibility}
The reason we identify the object closest to the extended-sets as a new extended-center is that we prove by Theorem \ref{the:theorem1} that such identification allows each extended-set to be within only one cluster. Therefore, after merging initial-clusters according to the extended-sets, each clustering center can share the representative degree of the extended-centers in the same cluster, so that its representative capability can cover the entire cluster, as shown in Figure \ref{fig:example}(B).

Since density is introduced in the formula (\ref{eq:dis}), the distance between extended-sets and high-density objects is shortened while the distance between extended-sets and low-density objects is increased. As a result, all of the identified extended-centers have high densities. Therefore, the formula (\ref{eq:dis}) allows ECAC to maximize representative capability with less extended-centers. As shown in Figure \ref{fig:example}(C), the blue object is the high-density neighbor of the clustering center (\emph{i.e.}, the red object), and the green object is the low-density neighbor of the clustering center. The number of objects within the blue shaded area is significantly more than the number of objects within the green shaded area. Obviously, choosing the blue object as a new extended-center will increase the representative capability of the clustering center more. 

\begin{theorem}
\label{the:theorem1}
In the non-overlapping dataset, identifying the high-density object closest to the extended-sets as a new extended-center can ensure that each extended-set is within only one cluster.
\end{theorem}
\begin{proof}
In the non-overlapping dataset, it is well known that the inter-cluster distance is greater than the intra-cluster distance. Suppose there exists an extended-set spanning different clusters, specifically, the extended-center $o_\dag$ in the extended-set $E_\psi$ is not in the cluster $cluster(c_\psi)$ but in the cluster $cluster(c_\divideontimes)$ ($c_\divideontimes$ is another clustering center).\\
$\because$ When $o_\dag$ is identified as a new extended-center, $o_\dag,E_\psi=\mathop{\arg\min}\limits_{o_i\in O-E, j\in\{1,2,\cdots,k\}}{dis(o_i,E_j)}$.\\
$\therefore$ For $\forall o_j\in cluster(c_\divideontimes)$, $dis(o_j,E_\divideontimes)>dis(o_\dag,E_\psi)$.\\
$\therefore$ For $\forall o_j\in cluster(c_\divideontimes)$, $\frac{\mathop{\min}\limits_{x\in E_\divideontimes}{\|o_j-x\|_2}}{\rho(o_j)}>\frac{\mathop{\min}\limits_{x\in E_\psi}{\|o_\dag-x\|_2}}{\rho(o_\dag)}$.\\
$\therefore$ When $\rho\left(o_j\right)>\rho(o_\dag)$, $\mathop{\min}\limits_{x\in E_\divideontimes}{\|o_j-x\|_2}>\mathop{\min}\limits_{x\in E_\psi}{\|o_\dag-x\|_2}$. Let's assume $o_\Theta=\mathop{\arg\min}\limits_{x\in E_\divideontimes}{\|o_j-x\|_2}$, and $o_\Phi=\mathop{\arg\min}\limits_{x\in E_\psi}{\|o_\dag-x\|_2}$.\\
$\therefore$  When $\rho\left(o_j\right)>\rho(o_\dag)$, $\|o_j-o_\Theta\|_2>\|o_\dag-o_\Phi\|_2$.\\
$\because$ $o_\dag\in\ cluster(c_\divideontimes)$, and $o_\Phi\in cluster(c_\psi)$.\\
$\therefore$ $\|o_j-o_\Theta\|_2>\|o_\dag-o_\Phi\|_2>\|cluster(c_\divideontimes)-cluster(c_\psi)\|_2$.\\
$\therefore$ $\|o_j-o_\Theta\|_2>\|cluster(c_\divideontimes)-cluster(c_\psi)\|_2$.\\
$\because$ Obviously, $\exists o_j\in cluster(c_\divideontimes)$, so that $o_j$ and $o_\Theta$ are neighbors.\\
$\therefore$ $\mathop{\max}\limits_{o_g\in cluster(c_\divideontimes)}{\|o_g-o_g^\star\|_2}>\|o_j-o_\Theta\|_2>\|cluster(c_\divideontimes)-cluster(c_\psi)\|_2$, in which $o_g^\star$ represents the neighbors of $o_g$.\\
$\therefore$ The conclusion that the intra-cluster distance is greater than the inter-cluster distance is obtained, which contradicts the objective fact that inter-cluster distance is greater than intra-cluster distance.\\
$\therefore$ The hypothesis does not hold, so the original proposition is proved.
\end{proof}

\begin{table}
\caption{The details of synthetic datasets.}
  \label{tab:synthetic}
  \begin{tabular}{cccc}
    \hline
    \multicolumn{1}{c}{ }&\multicolumn{1}{c}{Number}&\multicolumn{1}{c}{Clusters}&\multicolumn{1}{c}{Description}\\
    \hline
    \emph{unbalance} & 6,500 & 8 & density-unbalanced datasets\\
    \emph{unbalance2} & 6,500 & 8 & density-unbalanced datasets\\
    \emph{skewed} & 1,000 & 6 & density-unbalanced datasets\\
    \emph{asymmetric} & 1,000 & 5 & density-unbalanced datasets\\
    \emph{s1} & 5,000 & 15 & overlapping dataset\\
    \emph{s2} & 5,000 & 15 & overlapping dataset\\
    \emph{s3} & 5,000 & 15 & overlapping dataset\\
    \emph{s4} & 5,000 & 15 & overlapping dataset\\
    \emph{a1} & 3,000 & 20 & Gaussian dataset\\
    \emph{a2} & 5,250 & 35 & Gaussian dataset\\
    \emph{a3} & 7,500 & 50 & Gaussian dataset\\
    \emph{flame} & 240 & 2 & shaped dataset\\
    \emph{aggregation} & 788 & 7 & shaped dataset\\
    \emph{compound} & 399 & 6 & shaped dataset\\
    \emph{jain} & 373 & 2 & shaped dataset\\
    \emph{pathbased} & 300 & 3 & shaped dataset\\
    \emph{spiral} & 312 & 3 & shaped dataset\\
    \emph{t4.8k} & 8,000 & 6 & shaped dataset\\
    \emph{t7.10k} & 10,000 & 9 & shaped dataset\\
    \emph{t8.8k} & 8,000 & 8 & shaped dataset\\
    \hline
  \end{tabular}
\end{table}

\section{Experiments}
\label{sec:Experiments}

\subsection{Experimental settings}
In this subsection, we will describe the datasets, evaluation metrics, and center-based clustering algorithms to be optimized.

\subsubsection{Datasets}
We select some common synthetic and real-world datasets to test the proposed method. Synthetic datasets (from clustering basic benchmark \cite{franti2018k}) are
described in Table \ref{tab:synthetic}. Real-world datasets (from \emph{KEEL}-dataset repository \cite{derrac2015keel}) are described as follows:
\begin{itemize}
\item \textbf{\emph{Dermatology}} is a diagnostic dataset for erythemato-squamous diseases. Each object is the clinical evaluation of a patient, including 12 clinical features and 22 histopathological features. The patients are classified into 6 categories according to the degree of disease.

\item \textbf{\emph{Movement}} records 15 categories of hand movements in LIBRAS (Lingua BRAsileira de Sinais), each represented by 90 features.

\item \textbf{\emph{Optdigits}} is a dataset of 5,620 digits handwritten by 43 people, which are integers in the range of 0 to 16.

\item \textbf{\emph{Page}} contains 5,472 page layout blocks from 54 documents. These blocks are represented by 10 features and are divided into 5 categories: text, horizontal lines, graphics, vertical lines and images.

\item \textbf{\emph{Penbased}} is a dataset of 10,992 digits, each represented by 16 features, handwritten by 44 people.

\item \textbf{\emph{Texture}} records 5,500 texture objects, which can be divided into 11 categories, such as grass lawn, pressed calf leather, handmade paper, \emph{etc}.

\item \textbf{\emph{Satimage}} records the multispectral values of satellite image pixels in a 3*3 neighborhood, which can be divided into 7 categories, such as red soil, cotton crop, grey soil, \emph{etc}.

\item \textbf{\emph{Breast}} records 296 objects. They are described by 9 features, some of which are linear and some of which are nominal.

\item \textbf{\emph{Wifi\_loc}} records the Wi-Fi signal strength of 2,000 objects in 4 different rooms to predict which room each object is in.

\item \textbf{\emph{Banknote}} records some attributes of banknotes to determine whether they are counterfeit.
\end{itemize}

\subsubsection{Evaluation metrics}
We compare the accuracy of center-based clustering algorithms before and after they are optimized by ECAC. If the accuracy after being optimized by ECAC is higher than the original accuracy, then ECAC is effective. Two approaches are used to measure clustering accuracy:

\textbf{1)} We calculate the differences between the clustering result and the ground-truth label by NMI and RI indices,
\begin{align*}
NMI(U,V)=\frac{\sum_{i=1}^{|U|}{\sum_{j=1}^{|V|}{P(i,j)\log{\frac{P(i,j)}{P(i)P'(J)}}}}}{\sqrt{\sum_{i=1}^{\left|U\right|}{P(i)\log{P(i)}}\times\sum_{i=1}^{\left|V\right|}{P'(j)\log{P'(j)}}}}
\end{align*}
, in which $U$ is the ground-truth label, $V$ is the clustering result, $P(i)=|U_i|/N$, $P'(j)=|V_j|/N$, $P(i,j)=|V_j \cap U_i|/N$.
\begin{align*}
RI=\frac{TP+TN}{TP+FP+TN+FN}
\end{align*}
, in which $TP$ is the number of similar object pairs (\emph{i.e.}, their ground-truth labels are the same) being clustered into the same category, $TN$ is the number of dissimilar object pairs being clustered into different categories, $FP$ is the number of dissimilar object pairs being clustered into the same category, and $FN$ is the number of similar object pairs being clustered into different categories. The range of both NMI and RI is $\left[0,1\right]$. The closer the value is to 1, the more accurate the clustering result is.

\textbf{2)} We visualize the dataset and mark objects in different colors according to the clustering result. The more the number of objects marked in the same color in the same ground-truth cluster and the less the number of objects marked in the same color in different ground-truth clusters, the more accurate the clustering result is.

\subsubsection{Center-based clustering algorithms to be optimized}
In this paper, we choose three center-based clustering algorithms, DPC, K-means and Extreme clustering, to test the optimization effect of ECAC. Our choice does not imply that ECAC can only optimize them, and \textbf{we will discuss the generality of ECAC in Section \ref{sec:generality}}. The reason we choose them is that DPC and K-means are milestone center-based clustering algorithms (most existing center-based clustering algorithms are their improved versions) and Extreme clustering is recently proposed and cited by many researchers \cite{hao2021energy, zhu2021cdf, yang2022ecca, ma2022enhancing, zheng2021quickdsc,zhang2022novel}.

\begin{figure*}
  \centering
  \includegraphics[width=1.7in]{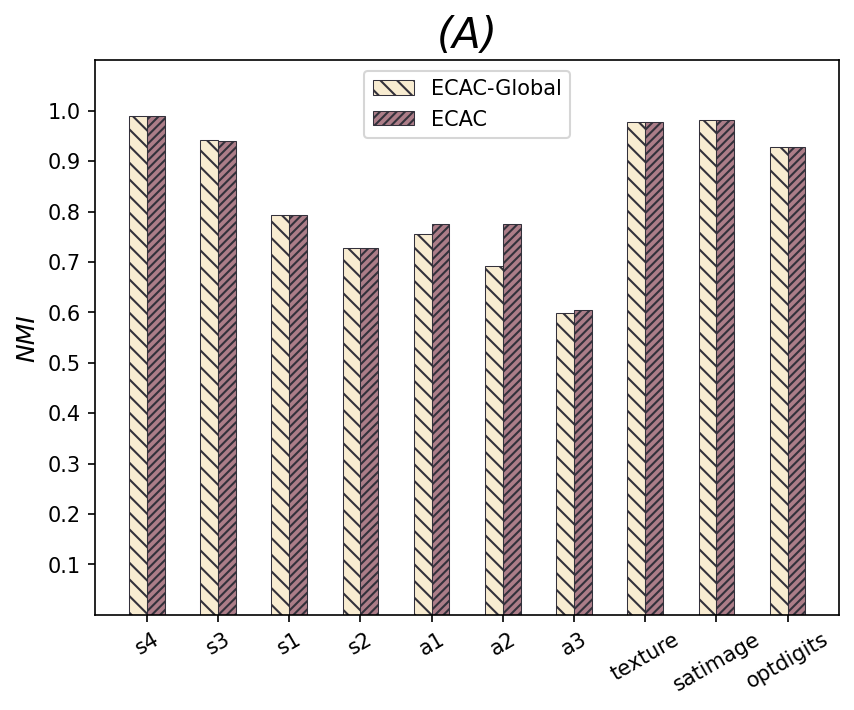}
  \includegraphics[width=1.7in]{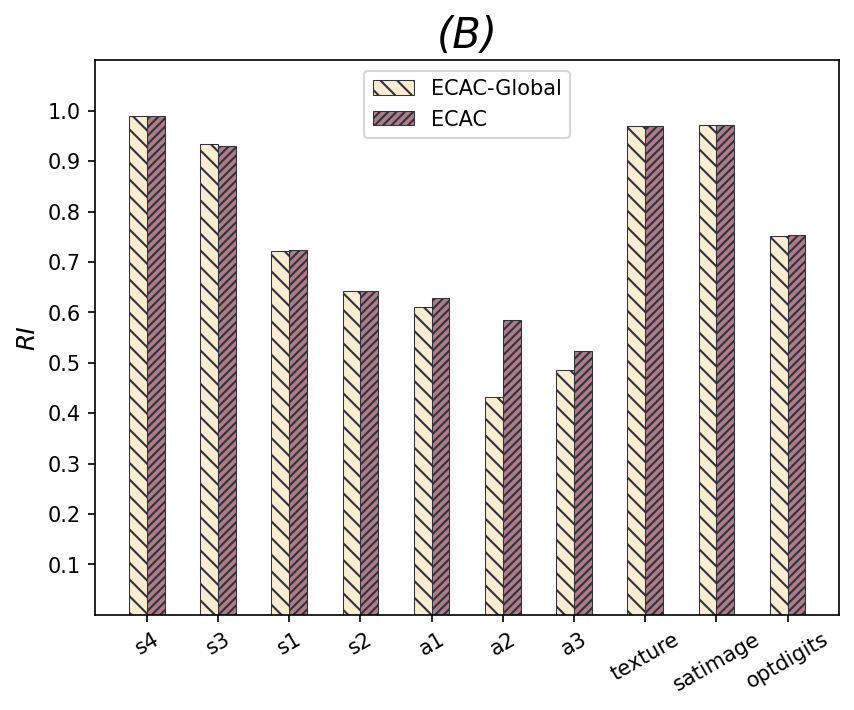}
  \includegraphics[width=1.7in]{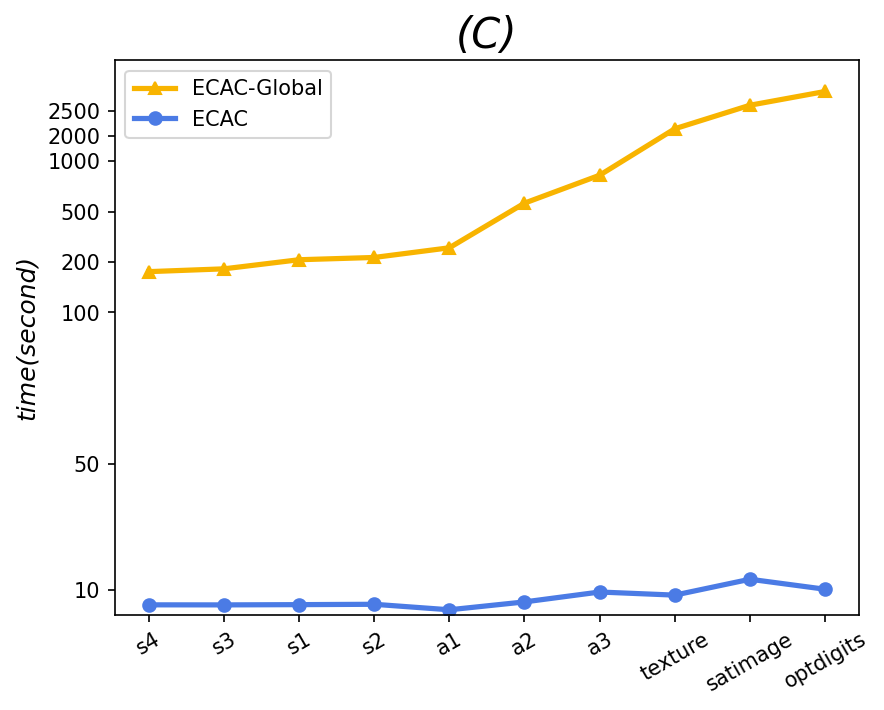}
  \caption{\textbf{ECAC \emph{vs.} ECAC-Global:} the accuracy of DPC is almost identical whether it is optimized by ECAC or by ECAC-Global, but the runtime of ECAC is much shorter than that of ECAC-Global. Obviously, narrowing the search for extended-centers is necessary and reasonable.}
  \label{fig:local}
\end{figure*}

\begin{figure}
  \centering
  \includegraphics[width=1.7in]{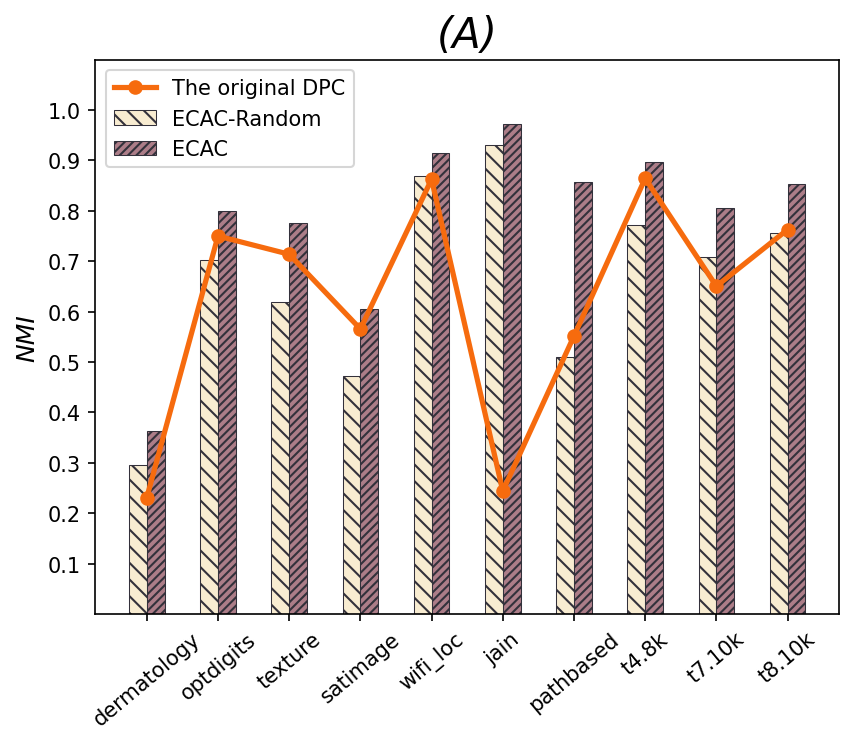}
  \includegraphics[width=1.7in]{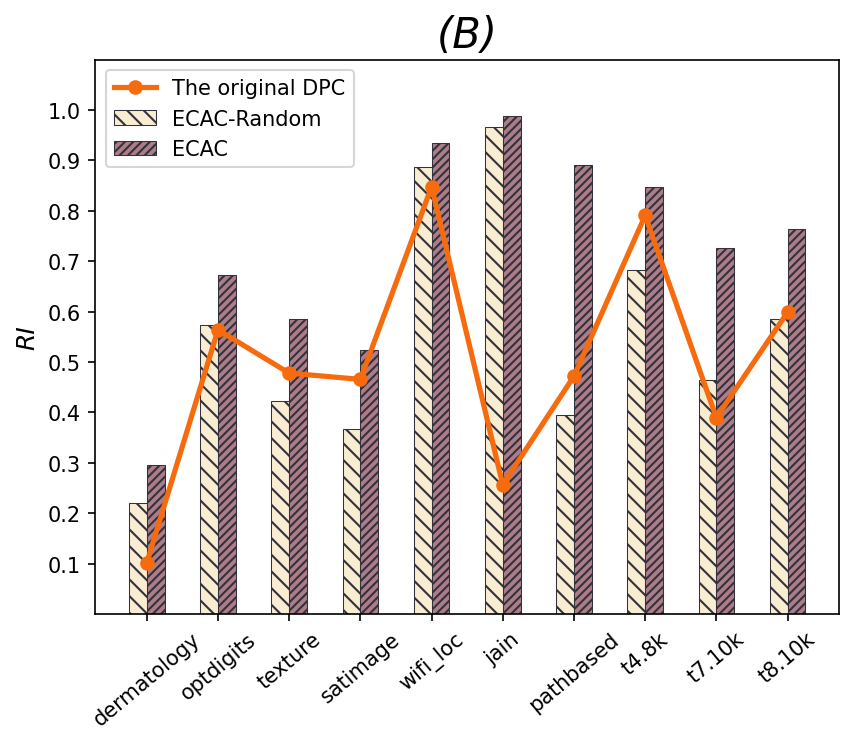}
  \caption{\textbf{ECAC \emph{vs.} ECAC-Random:} ECAC-Random reduces the accuracy of DPC on many datasets. ECAC effectively improves the accuracy of DPC on all datasets. Therefore, it is an effective strategy to identify the high-density object closest to the extended-sets as a new extended-center.}
  \label{fig:ECAC strategy}
\end{figure}

\begin{figure}
  \centering
  \includegraphics[width=1.7in]{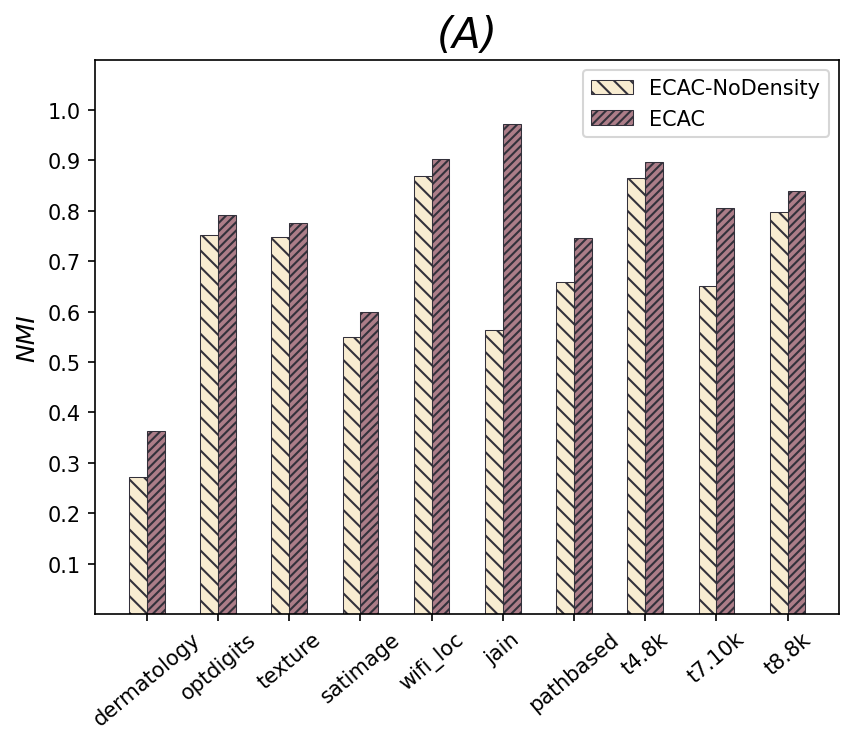}
  \includegraphics[width=1.7in]{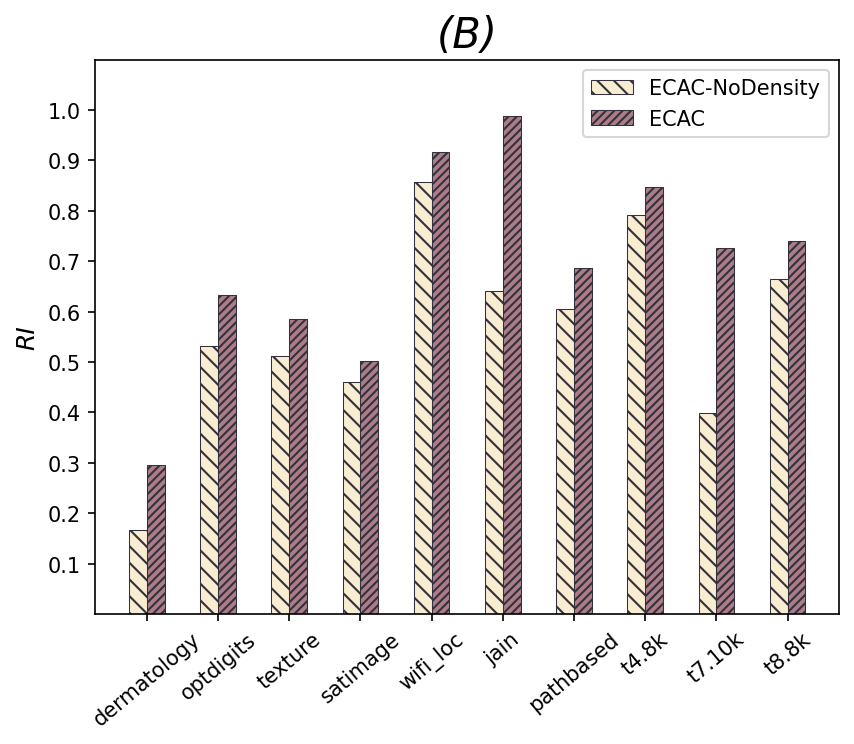}
  \caption{\textbf{ECAC \emph{vs.} ECAC-NoDensity:} the accuracy of the DPC optimized by ECAC is higher than that of the DPC optimized by ECAC-NoDensity on all datasets. Obviously, the formula (\ref{eq:dis}) with density can maximize the representative capability of each extended-center.}
  \label{fig:density}
\end{figure}

\subsection{Ablation experiments}
In this subsection, we will verify the necessity of the components of ECAC.

\subsubsection{The local search}
\label{sec:local}
We propose to search for new extended-centers only within the neighborhood of identified extended-centers with radius $2\delta$. The reason is that the objects around high-density objects (we have discussed in Section \ref{sec:Feasibility} that identified extended-centers are all high-density objects) are most likely to have high densities. Here, we verify the feasibility of this proposal in terms of accuracy and runtime. We refer to the ECAC with global search as ECAC-Global. We compare ECAC with ECAC-Global on \emph{optdigits}, \emph{satimage}, \emph{texture}, \emph{a1}, \emph{a2}, \emph{a3}, \emph{s1}, \emph{s2}, \emph{s3}, and \emph{s4} datasets. Figure \ref{fig:local} shows the accuracy (NMI as well as RI) of the DPC after being optimized by ECAC and ECAC-Global, as well as the runtimes of ECAC and ECAC-Global. The experimental results show that the accuracy of DPC is almost identical whether it is optimized by ECAC or by ECAC-Global, indicating that ECAC and ECAC-Global have similar optimization abilities. However, the runtime of ECAC is much shorter than that of ECAC-Global. ECAC-Global runs for more than 150 seconds on all datasets, which is unacceptable. Especially on \emph{optdigits}, ECAC-Global runs for 2,881.68 seconds, but ECAC only runs for 10.09 seconds. Therefore, the proposal to search for new extended-centers only within the neighborhood of identified extended-centers with radius $2\delta$ is reasonable and necessary.

\subsubsection{The ECAC strategy}
As mentioned in Section \ref{sec:extended-centers}, we identify the high-density object closest to the extended-sets as a new extended-center (Hereafter referred to as the ECAC strategy). To verify the effectiveness of the ECAC strategy, we compare it with the random strategy on 10 datasets. These datasets are \emph{dermatology}, \emph{optdigits}, \emph{texture}, \emph{satimage}, \emph{wifi\_loc}, \emph{jain}, \emph{pathbased}, \emph{t4.8k}, \emph{t7.10k}, and \emph{t8.8k}. The random strategy randomly selects objects as extended-centers and then assigns them to the extended-set where the nearest clustering center is located. Here, we refer to the ECAC with the random strategy as ECAC-Random. Figure \ref{fig:ECAC strategy} shows the accuracy (NMI as well as RI) of the original DPC and the accuracy of the DPC after being optimized by ECAC and ECAC-Random. The experimental results show that ECAC-Random instead reduces the accuracy of DPC on \emph{optdigits}, \emph{texture}, \emph{satimage}, \emph{wifi\_loc}, \emph{pathbased}, and \emph{t4.8k}. We believe the main reason is that the random strategy does not guarantee that each extended-set is only within one cluster, resulting in objects in different clusters being identified as a category. On other datasets, the optimization of ECAC-Random is also inferior to that of ECAC. we believe that the main reason is that the random strategy is unable to spread extended-centers uniformly throughout cluster, resulting in the representative capability not covering the entire cluster. As for ECAC, it effectively improves the accuracy (NMI as well as RI) of DPC on all datasets. Therefore, it is an effective strategy to identify the high-density object closest to the extended-sets as a new extended-center.

\begin{figure*}
  \centering
  \includegraphics[width=1.5in]{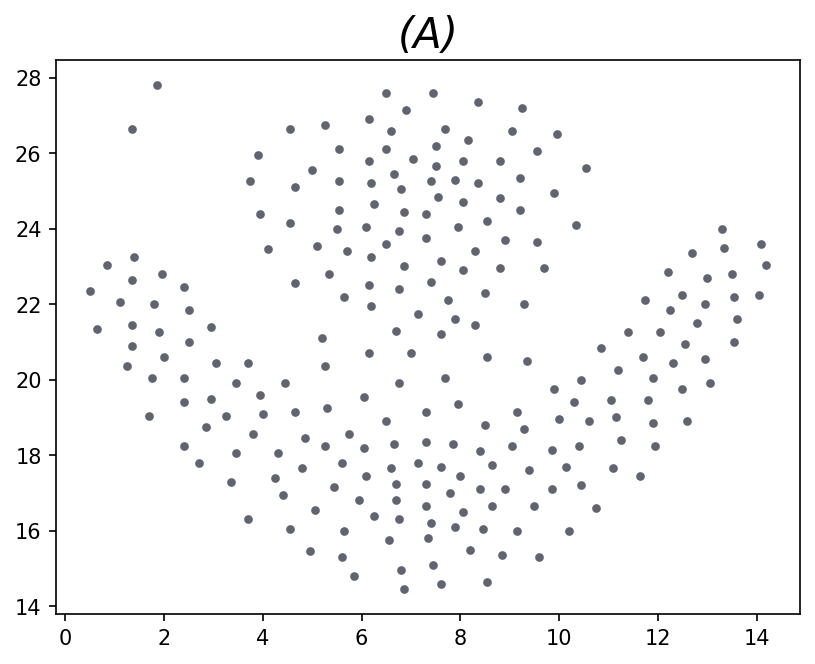}
  \includegraphics[width=1.5in]{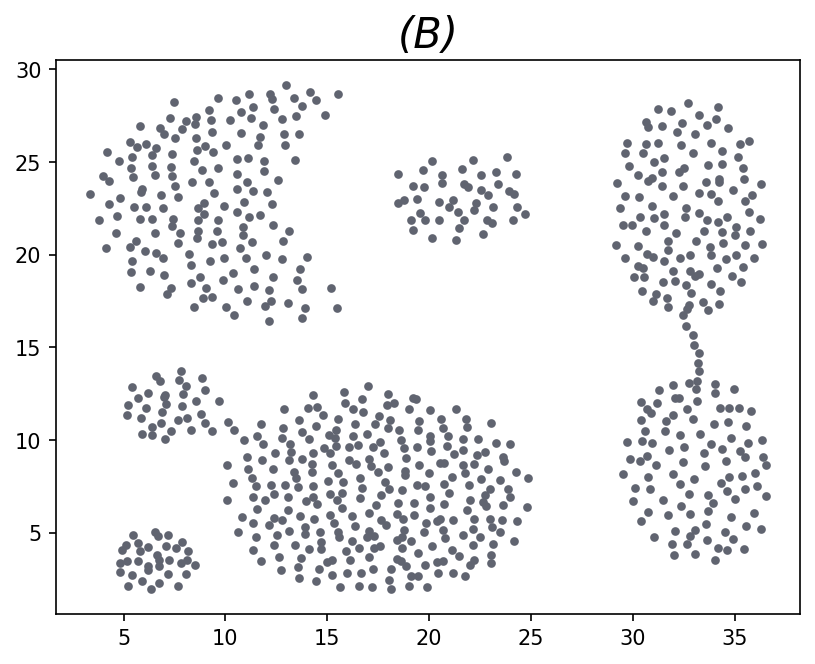}
  \includegraphics[width=1.5in]{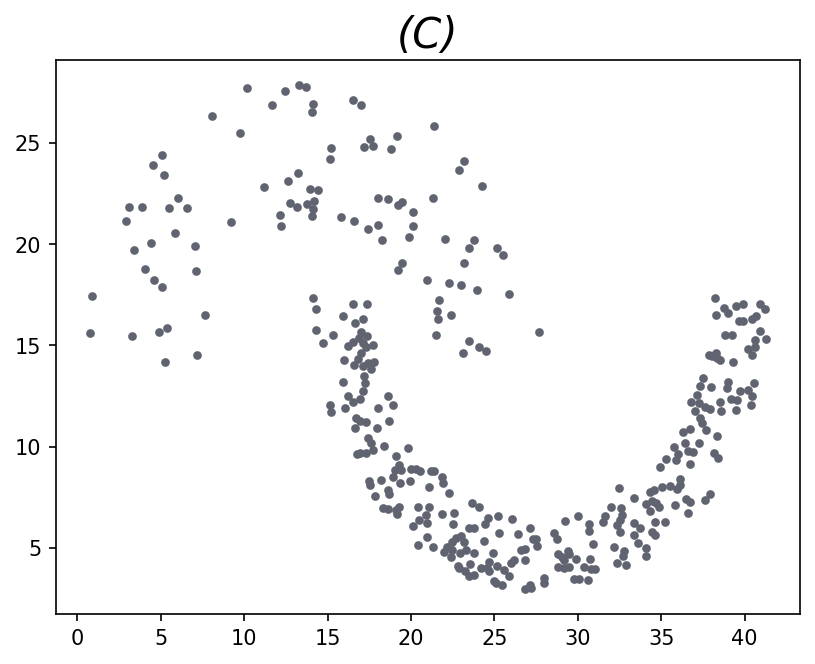}
  \includegraphics[width=1.5in]{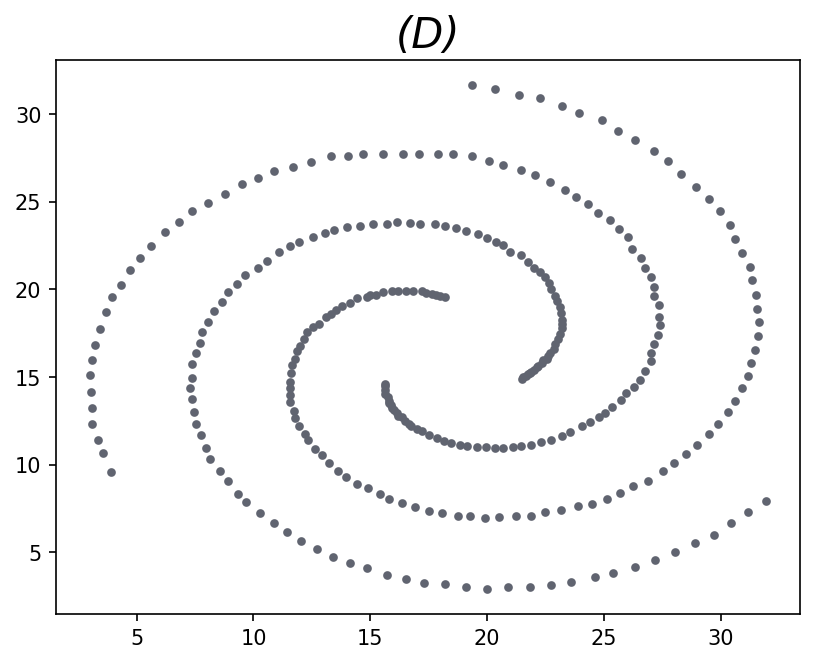}\\
  \includegraphics[width=1.5in]{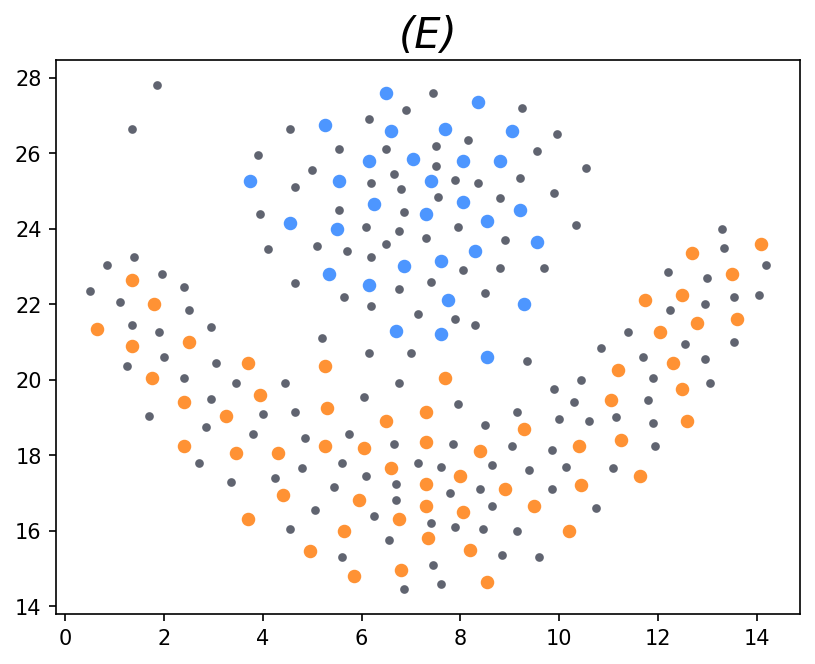}
  \includegraphics[width=1.5in]{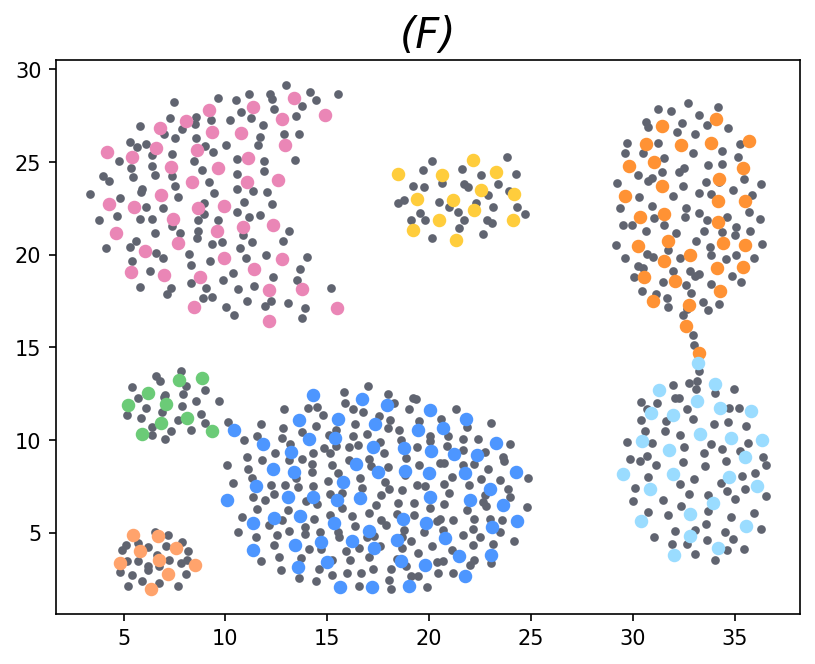}
  \includegraphics[width=1.5in]{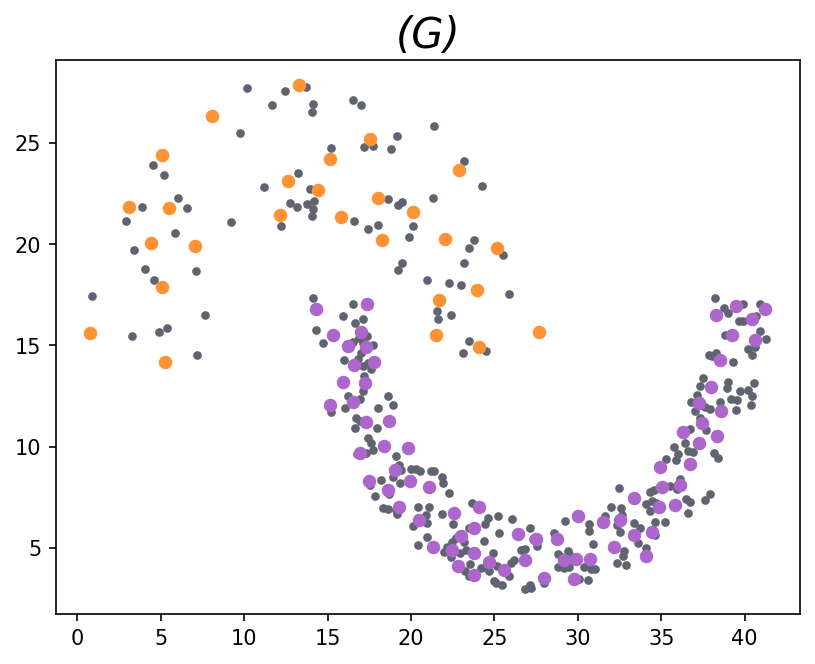}
  \includegraphics[width=1.5in]{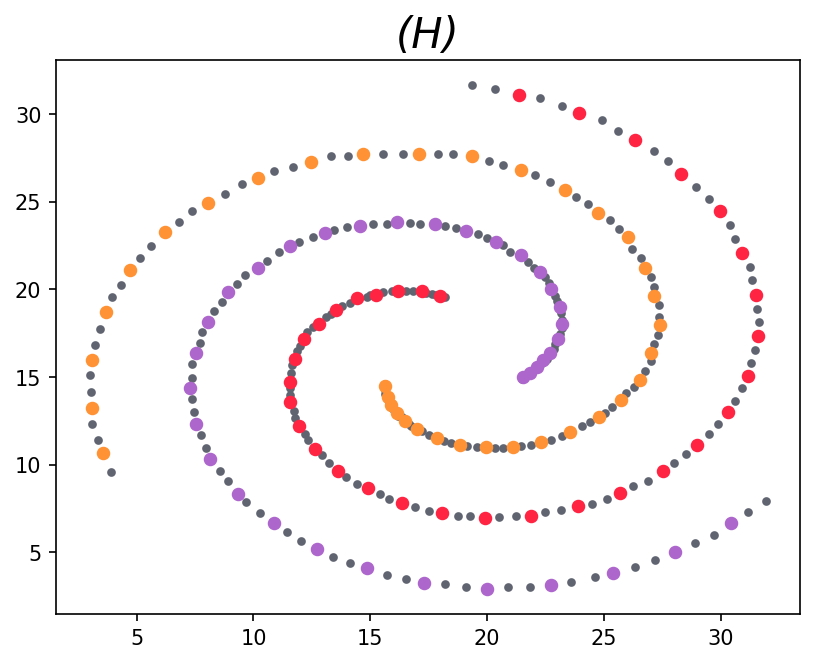}
  \caption{\textbf{The distribution of shaped datasets (A, B, C, D) and the extended-centers identified by ECAC (E, F, G, H):} Different extended-sets are marked in different colors. The extended-centers marked in the same color are in the same cluster, and the extended-sets inherit the cluster shape. Therefore, ECAC is robust to shaped datasets.}
  \label{fig:shape}
\end{figure*}

\begin{figure*}
  \centering
  \includegraphics[width=1.5in]{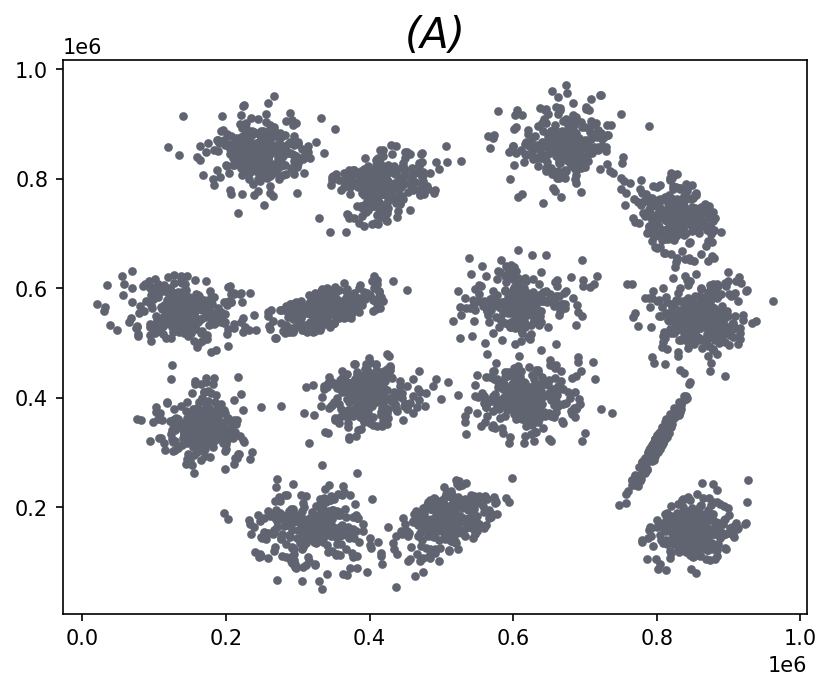}
 \includegraphics[width=1.5in]{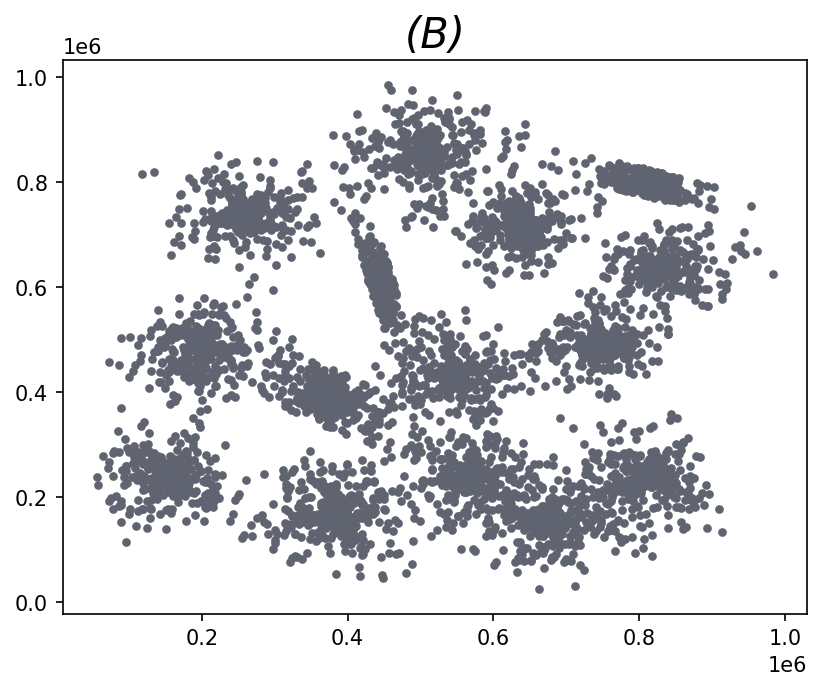}
 \includegraphics[width=1.5in]{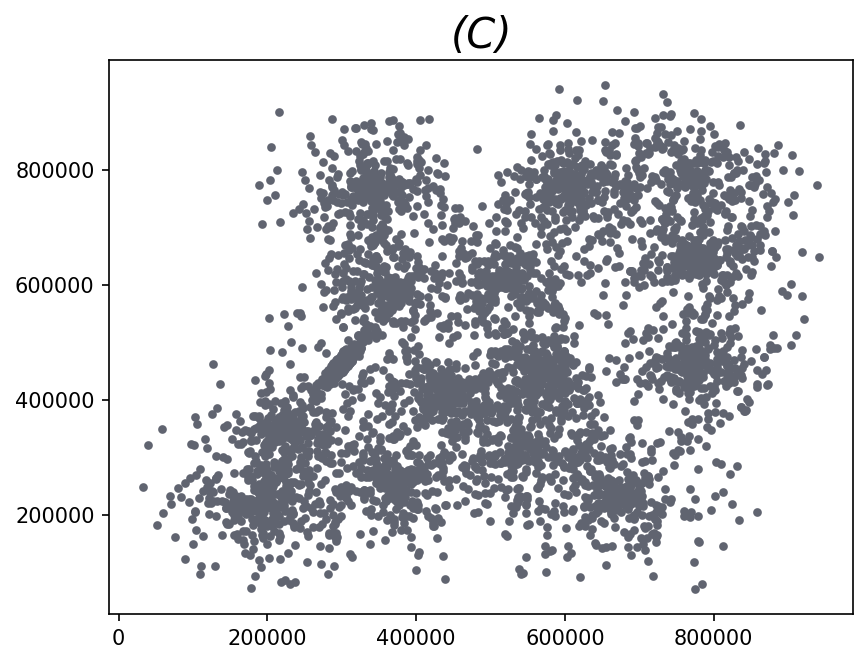}
 \includegraphics[width=1.5in]{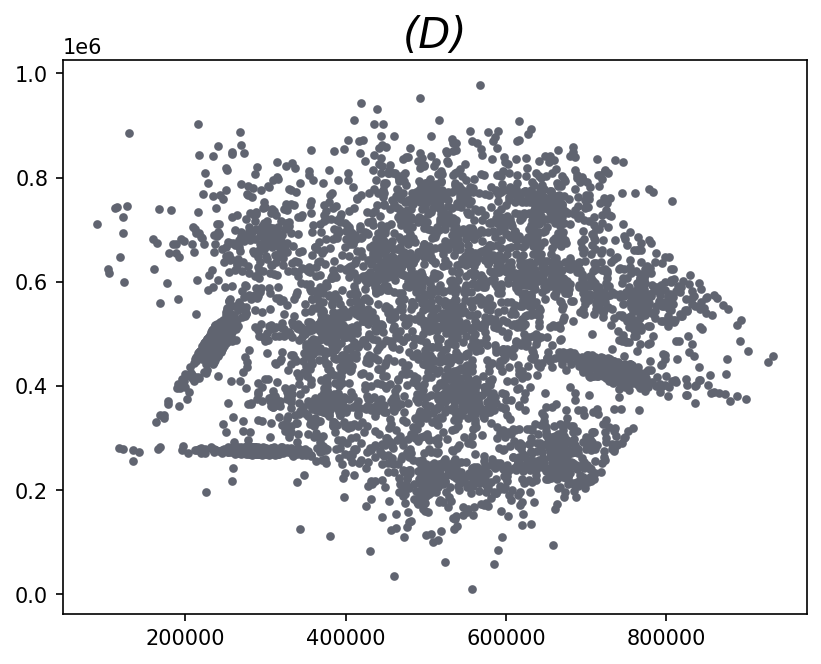}\\
 \includegraphics[width=1.5in]{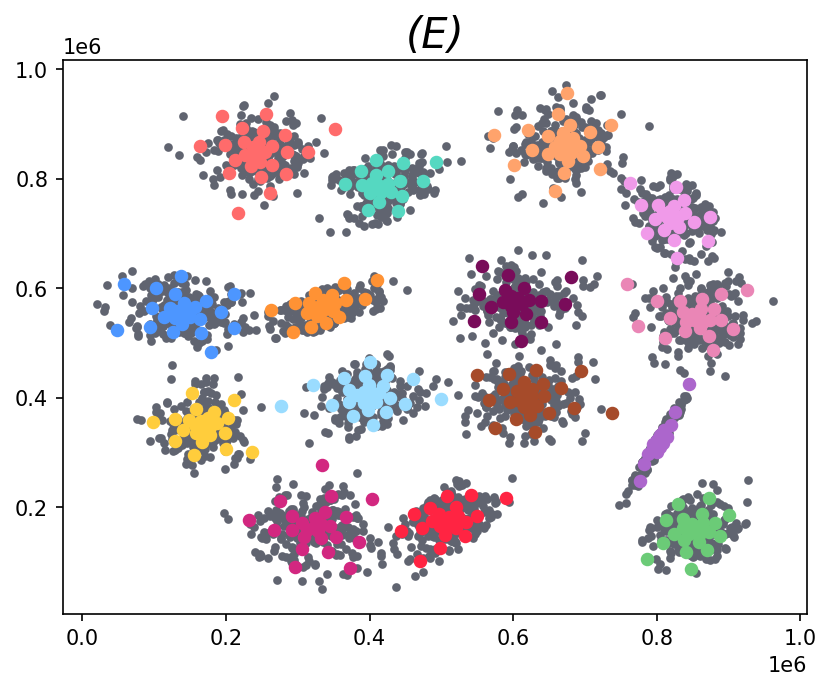}
 \includegraphics[width=1.5in]{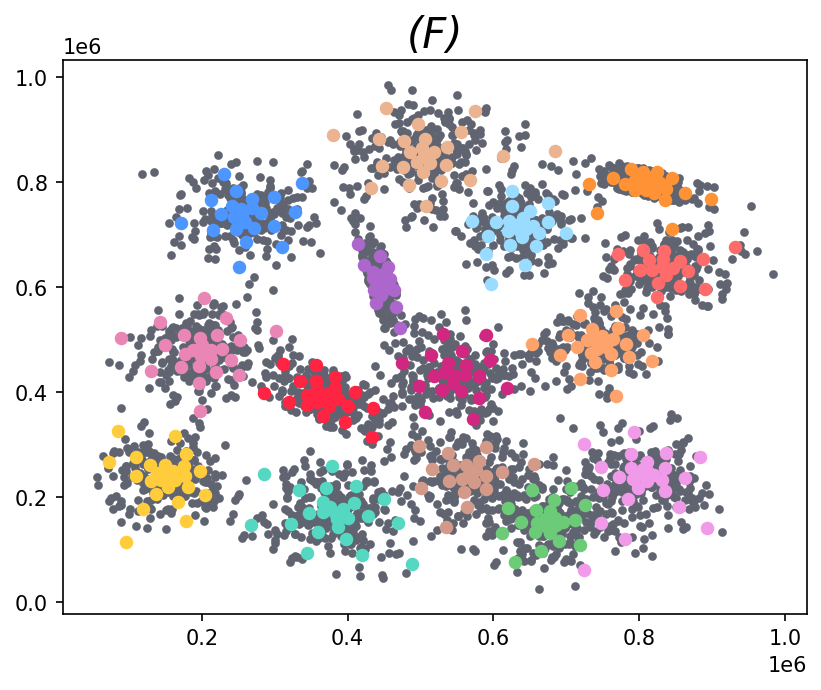}
 \includegraphics[width=1.5in]{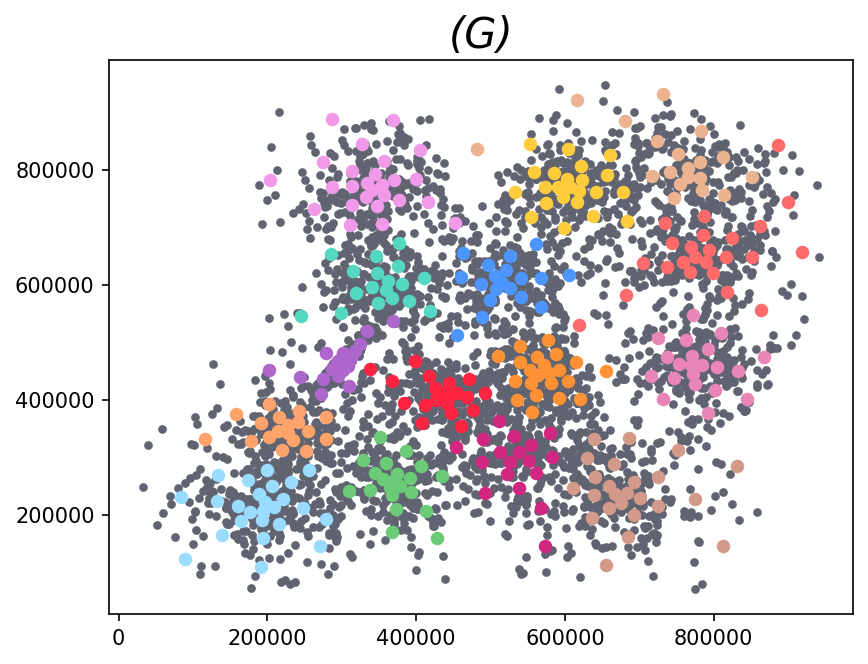}
 \includegraphics[width=1.5in]{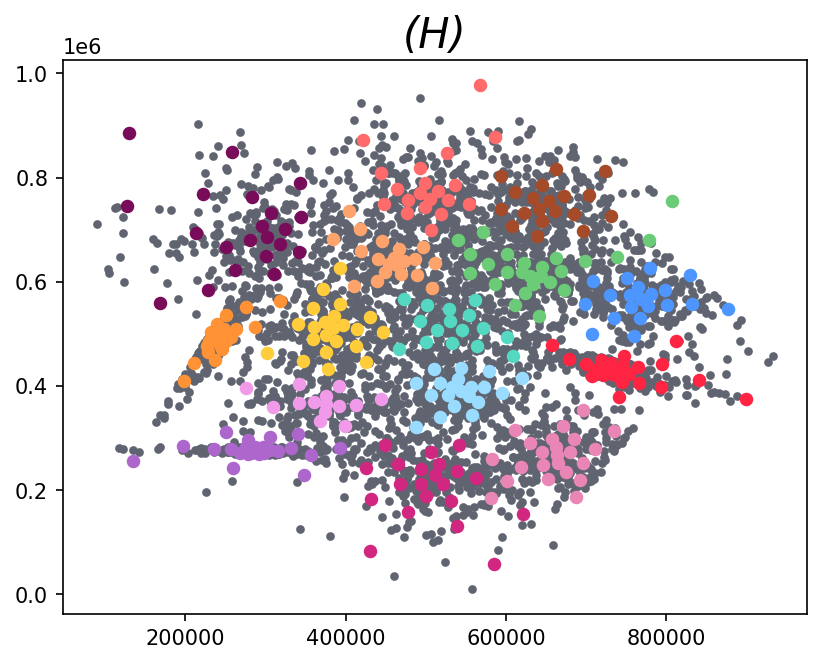}
  \caption{\textbf{The distribution of overlapping datasets (A, B, C, D) and the extended-centers identified by ECAC (E, F, G, H):} Different extended-sets are marked in different colors. No matter how overlapping clusters are, the extended-centers in the same cluster are marked in only one color. After marking, the outline of clusters in \emph{s4} becomes visible. Therefore, ECAC is also robust to overlapping datasets.}
  \label{fig:overlapping}
\end{figure*}

\subsubsection{The density}
\label{sec:density}
We mentioned in Section \ref{sec:Feasibility} that the formula (\ref{eq:dis}) with density can maximize the representative capability of each extended-center. Here, we will experimentally verify our judgment. We refer to the ECAC without density as ECAC-NoDensity. To objectively compare the average representative capability of the extended-centers identified by ECAC and ECAC-NoDensity, we terminate ECAC and ECAC-NoDensity when they identify at least the same number of extended-centers in the same cluster. Figure \ref{fig:density} shows the accuracy (NMI as well as RI) of the DPC after being optimized by ECAC and ECAC-NoDensity on \emph{dermatology}, \emph{optdigits}, \emph{texture}, \emph{satimage}, \emph{wifi\_loc}, \emph{jain}, \emph{pathbased}, \emph{t4.8k}, \emph{t7.10k}, and \emph{t8.8k} datasets. The experimental results show that the accuracy of the DPC optimized by ECAC is higher than that of the DPC optimized by ECAC-NoDensity on all datasets. Especially on \emph{jain}, the accuracy of the DPC optimized by ECAC is almost double that of the DPC optimized by ECAC-NoDensity. Obviously, the average representative capability of the extended-centers identified by ECAC is significantly higher than that of the extended-centers identified by ECAC-NoDensity.

\subsection{Robustness experiments}
In this subsection, we will verify the robustness of ECAC to diverse datasets and its generality to different center-based clustering algorithms.

\subsubsection{Robustness to datasets}
Clustering is an unsupervised analysis task. Before clustering, the dataset to be analyzed is unknown. Therefore, ECAC is a practical optimization method only if it is robust to diverse datasets. Here, we will test the effectiveness of ECAC on different datasets. Specifically, we designate an object in each cluster as a clustering center to verify whether the extended-centers derived from the same clustering center (\emph{i.e.}, an extended-set) are in the same cluster, and whether they are spread evenly across the entire cluster.

\begin{figure*}
  \centering
  \includegraphics[width=1.5in]{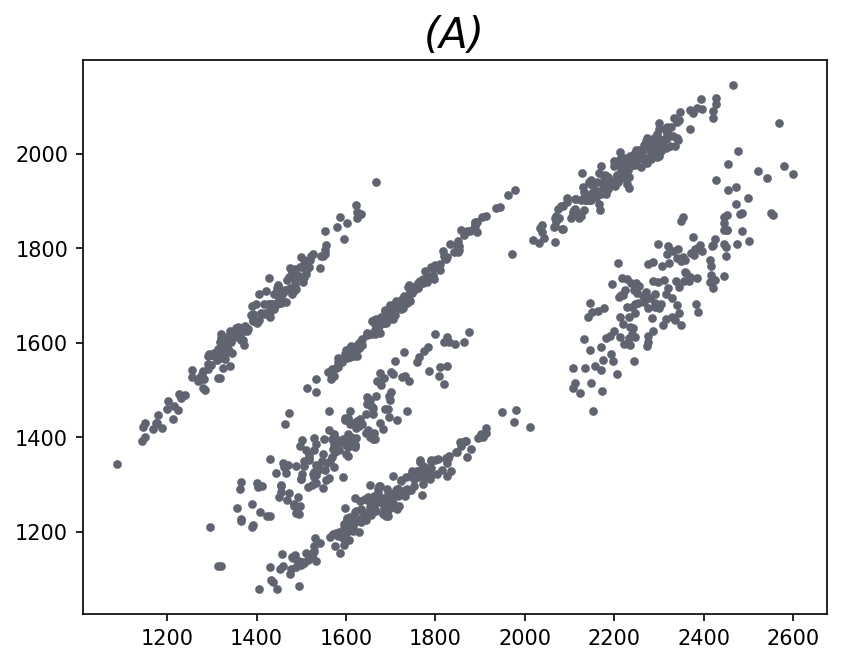}
  \includegraphics[width=1.5in]{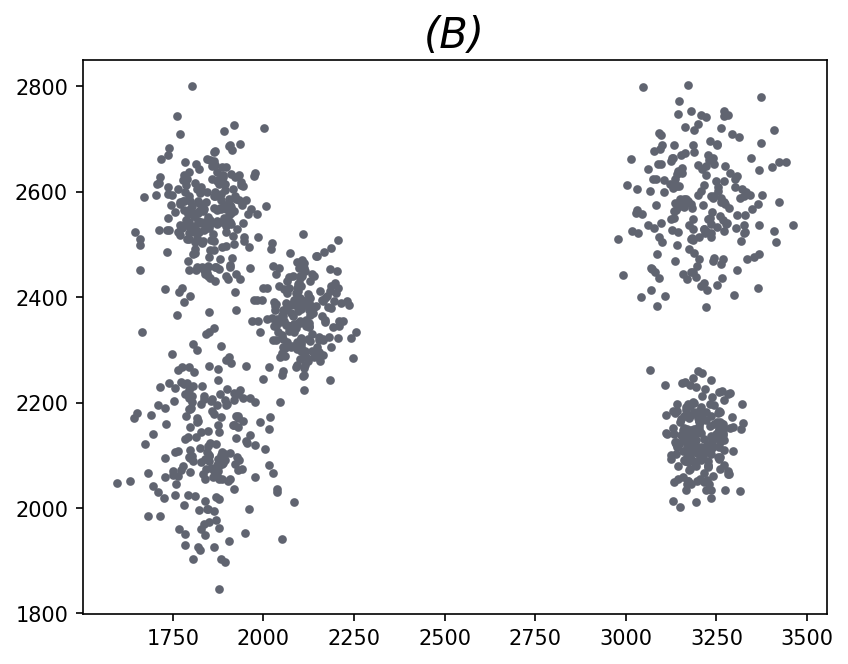}
  \includegraphics[width=1.5in]{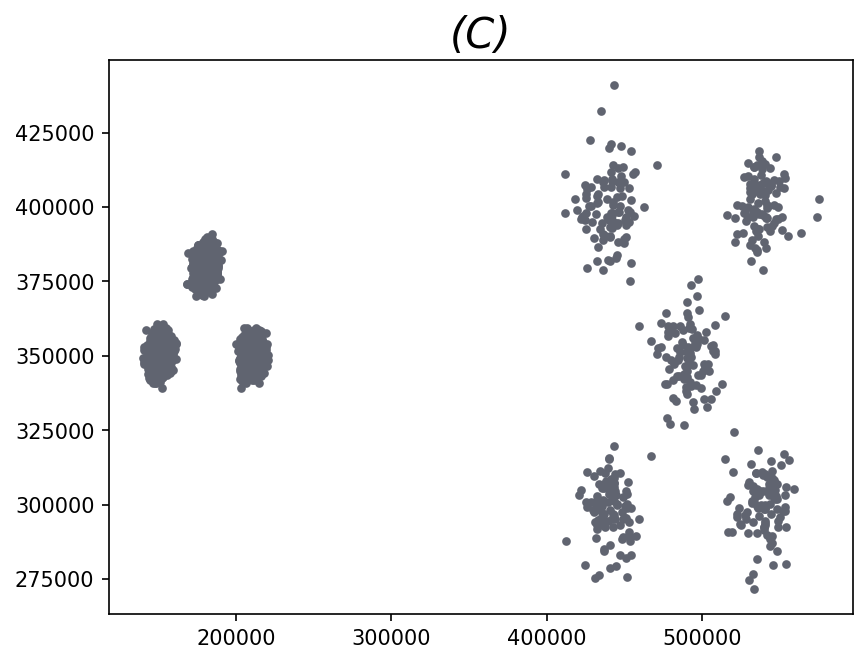}
  \includegraphics[width=1.5in]{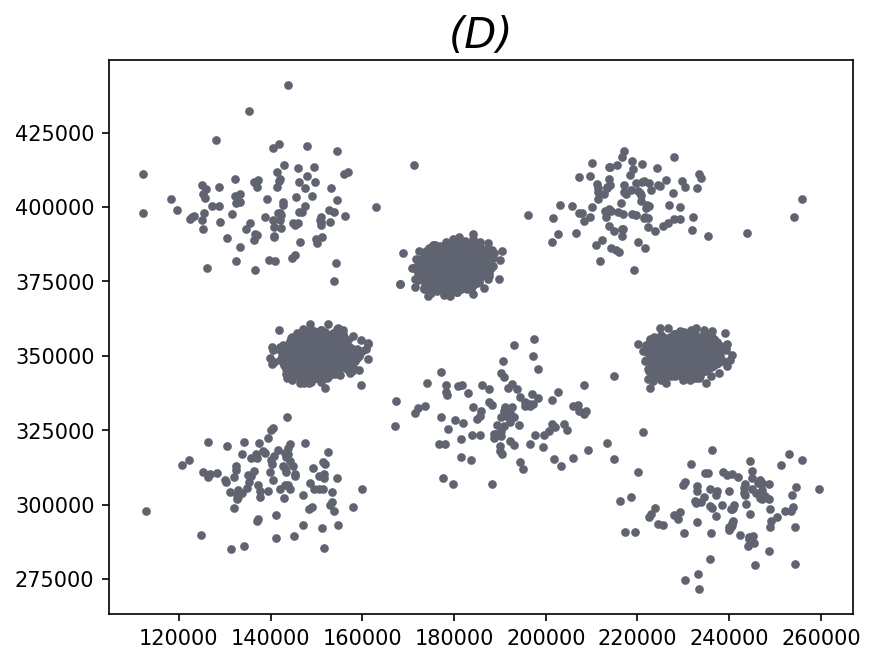}\\
  \includegraphics[width=1.5in]{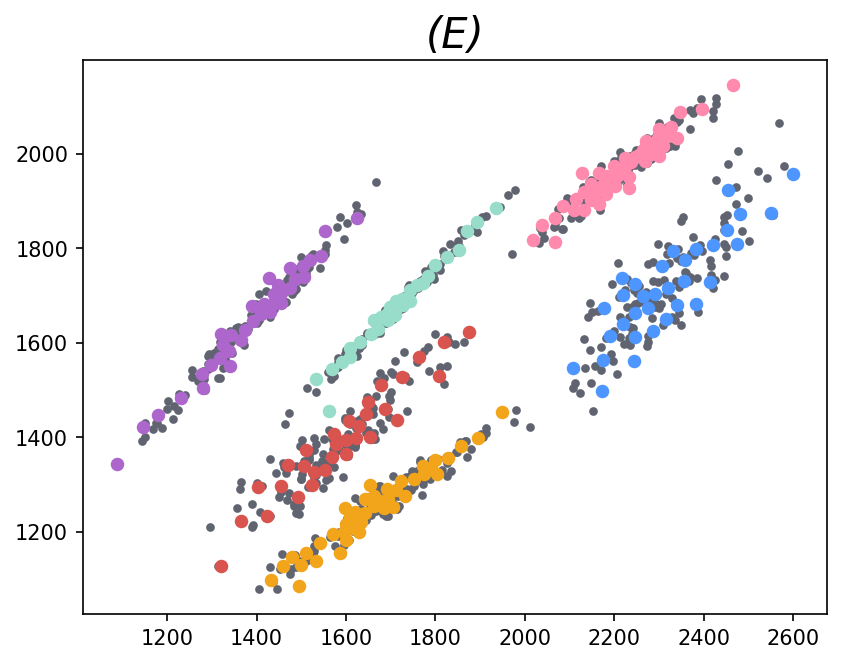}
  \includegraphics[width=1.5in]{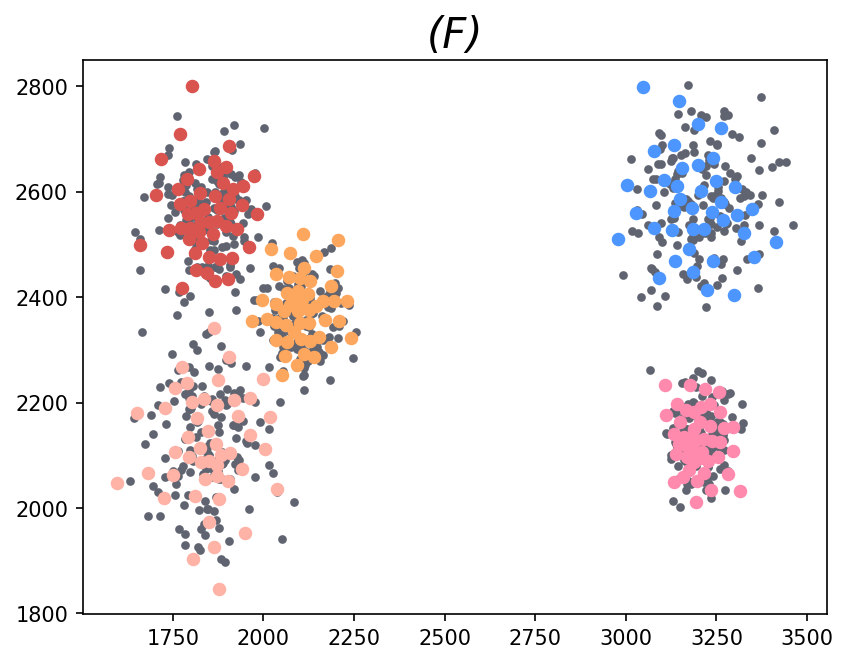}
  \includegraphics[width=1.5in]{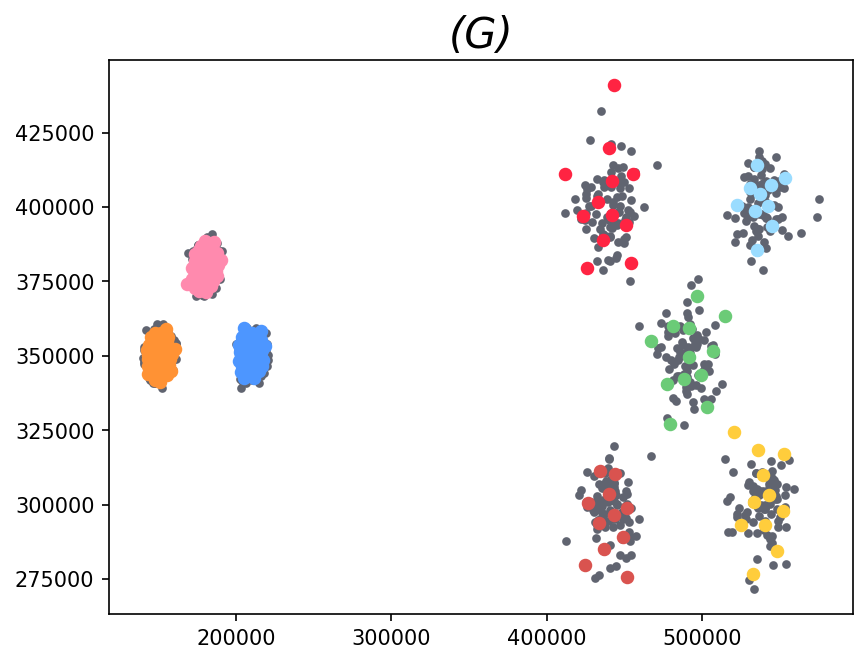}
  \includegraphics[width=1.5in]{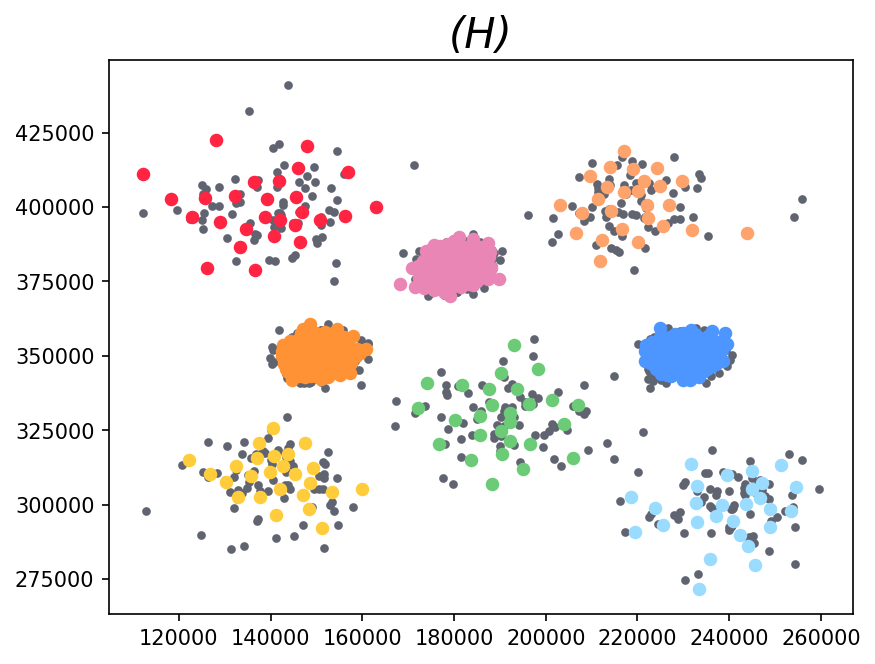}
  \caption{\textbf{The distribution of density-unbalanced datasets (A, B, C, D) and the extended-centers identified by ECAC (E, F, G, H):} Different extended-sets are marked in different colors. Regardless of the density differences between clusters, ECAC always correctly identifies only one extended-set within each cluster. ECAC is robust to density-unbalanced datasets.}
  \label{fig:unbalance}
\end{figure*}

\textbf{\emph{The robustness to shaped datasets:}} Traditional center-based clustering algorithms usually fail to identify shaped clusters because shaped clusters do not have obvious distribution trends. Therefore, it is necessary to verify whether ECAC is robust to shaped datasets. We test ECAC on 4 shaped datasets, \emph{i.e.}, \emph{flame}, \emph{aggregation}, \emph{jain}, and \emph{spiral}. In \emph{flame}, the upper and lower clusters are \emph{flame}-like. In \emph{jain}, clusters are crescent-shaped. In \emph{spiral}, 3 curve clusters are spirally intertwined. In \emph{aggregation}, clusters have various shapes, and 2 clusters are connected. Figure \ref{fig:shape} shows their distribution (row 1) and the extended-centers identified by ECAC (row 2), where different extended-sets are marked in different colors. The experimental results show that ECAC successfully identifies extended-centers along the shape of the clusters. These extended-centers are spread evenly across each cluster, so extended-sets inherit the shape of the clusters. Clearly, ECAC is robust to the shaped datasets.

\textbf{\emph{The robustness to overlapping datasets:}} Theorem \ref{the:theorem1} theoretically ensures ECAC is valid for non-overlapping datasets. Is ECAC also valid for overlapping datasets? Here, we test ECAC on S-sets. S-sets contains 4 overlapping datasets, \emph{s1}, \emph{s2}, \emph{s3}, and \emph{s4}, each of which consists of 15 Gaussian clusters. From \emph{s1} to \emph{s4}, the clusters overlap more and more, and even the cluster boundaries of \emph{s4} are indistinguishable by the naked eye. Figure \ref{fig:overlapping} shows their distribution (row 1) and the extended-centers identified by ECAC (row 2), where different extended-sets are marked in different colors. The experimental results show that no matter how overlapping clusters are, the extended-centers in the same cluster are marked in only one color, and the extended-centers in different clusters are marked in different colors. After marking, the outline of clusters in \emph{s4} becomes visible. In conclusion, ECAC is not only valid for non-overlapping datasets, but also some overlapping datasets.

\begin{figure}
  \centering
  \includegraphics[width=1.7in]{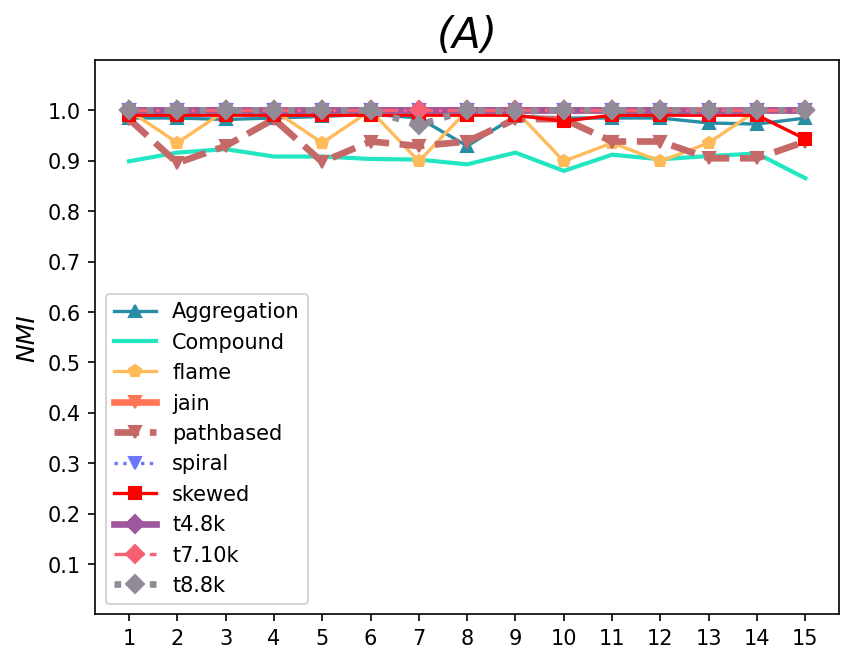}
  \includegraphics[width=1.7in]{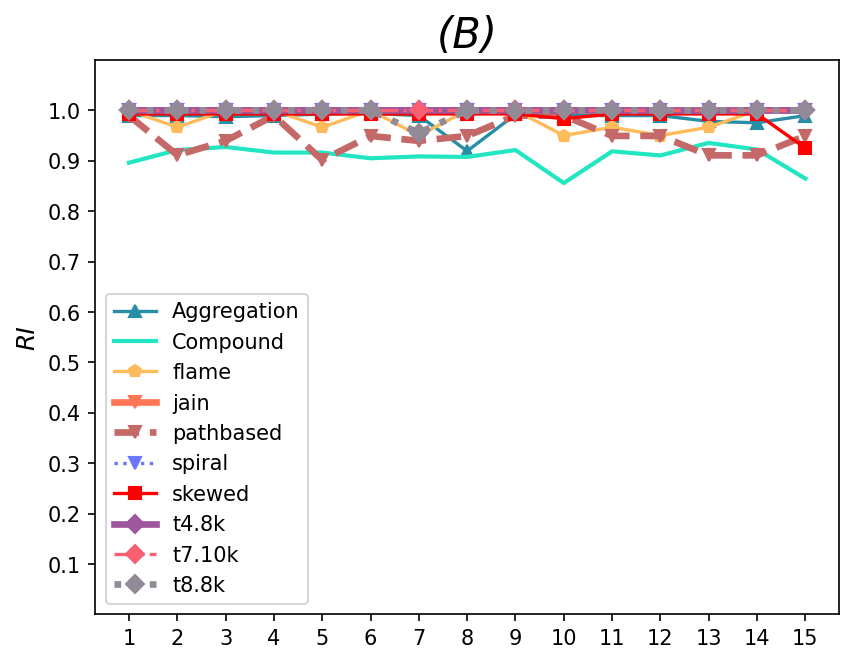}
  \caption{\textbf{The optimization effect of ECAC with different objects as clustering centers:} Regardless of which objects serve as clustering centers, the optimization effect of ECAC is very stable. Therefore, ECAC is a general method and can optimize different center-based clustering algorithms.}
  \label{fig:centers}
\end{figure}

\textbf{\emph{The robustness to density-unbalanced datasets:}} ECAC identifies the object with the smallest distance-to-density ratio (see the formula (\ref{eq:dis}) for details) as a new extended-center. In the density-unbalanced dataset, objects in sparse clusters are not only less dense but also farther away from other objects, which will inevitably lead to the identification order of extended-centers being biased towards dense clusters. So, is ECAC valid for density-unbalanced datasets? Here, we test ECAC on 4 density-unbalanced datasets, namely \emph{skewed}, \emph{asymmetric}, \emph{unbalance}, and \emph{unbalance2}. The density difference ratios (\emph{i.e.}, the ratio of the average density of objects in the densest cluster to the average density of objects in the sparsest cluster) of \emph{skewed} and \emph{asymmetric} are 2.85, and 3.04, respectively. The density difference ratios of \emph{unbalance2} and \emph{unbalance} are much higher than those of the first two datasets, reaching 87 and 85.57, respectively. Figure \ref{fig:unbalance} shows their distribution (row 1) and the extended-centers identified by ECAC (row 2), where different extended-sets are marked in different colors. The experimental results show that the identification order of extended-centers does not affect ECAC. Regardless of the density differences between clusters, ECAC always correctly identifies only one extended-set within each cluster. ECAC is robust to density-unbalanced datasets.

\begin{figure*}
  \centering
  \includegraphics[width=1.1in]{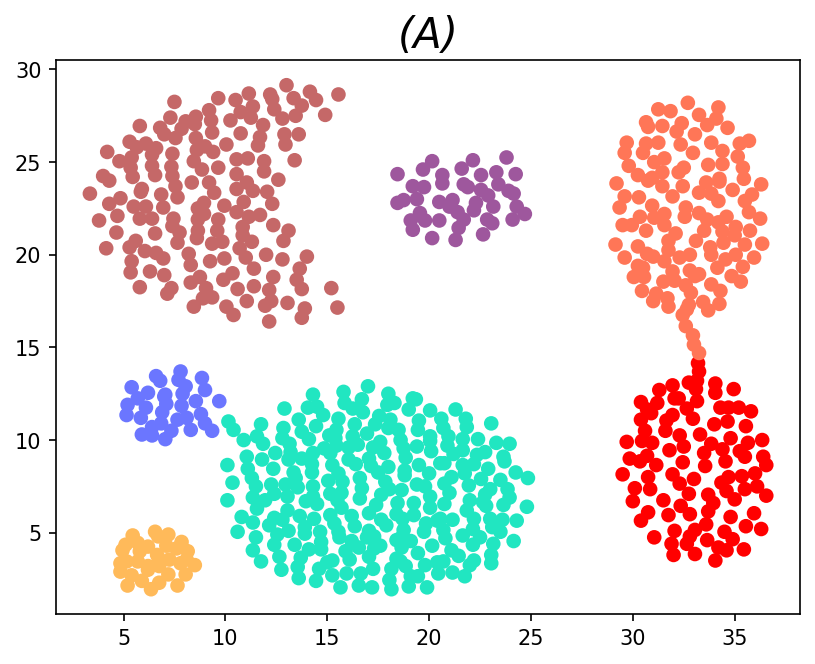}
  \includegraphics[width=1.1in]{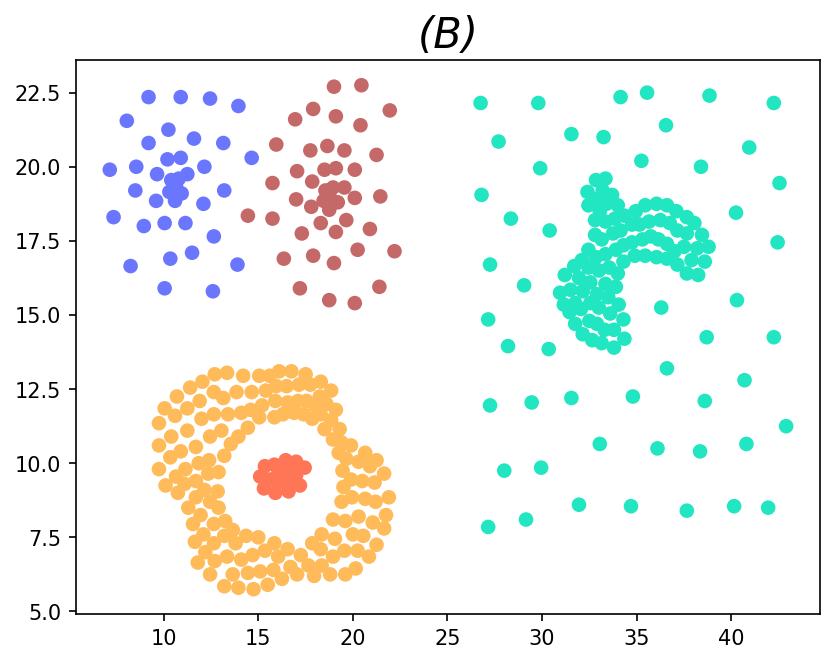}
  \includegraphics[width=1.1in]{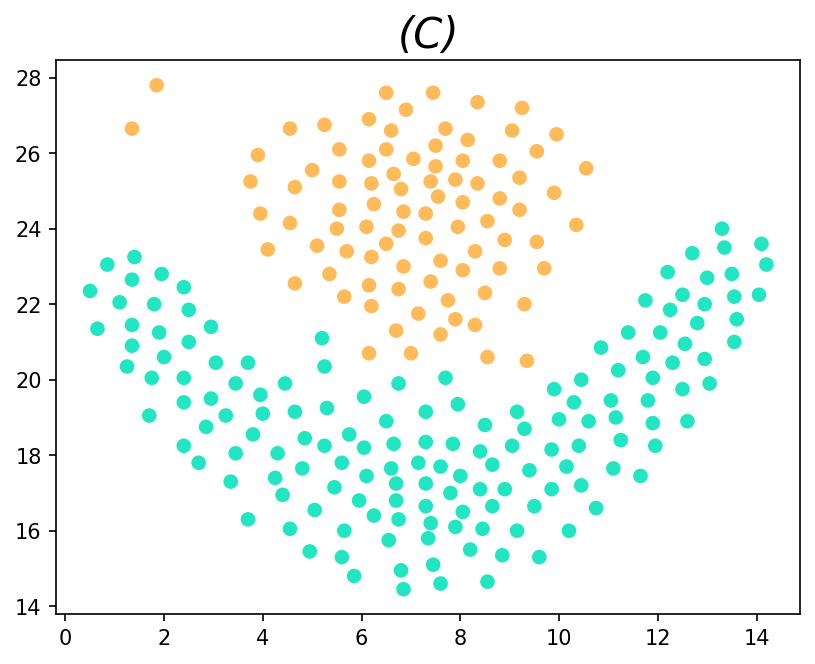}
  \includegraphics[width=1.1in]{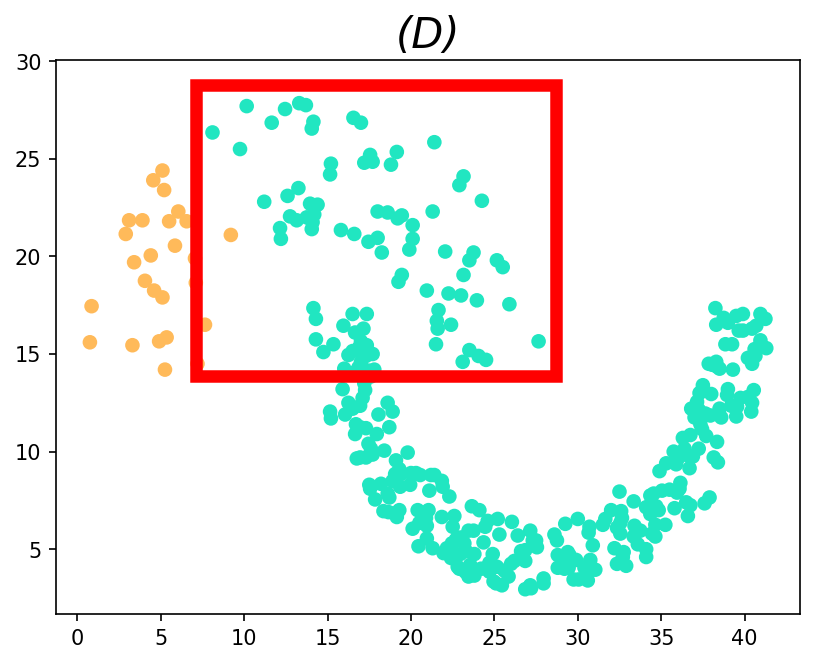}
  \includegraphics[width=1.1in]{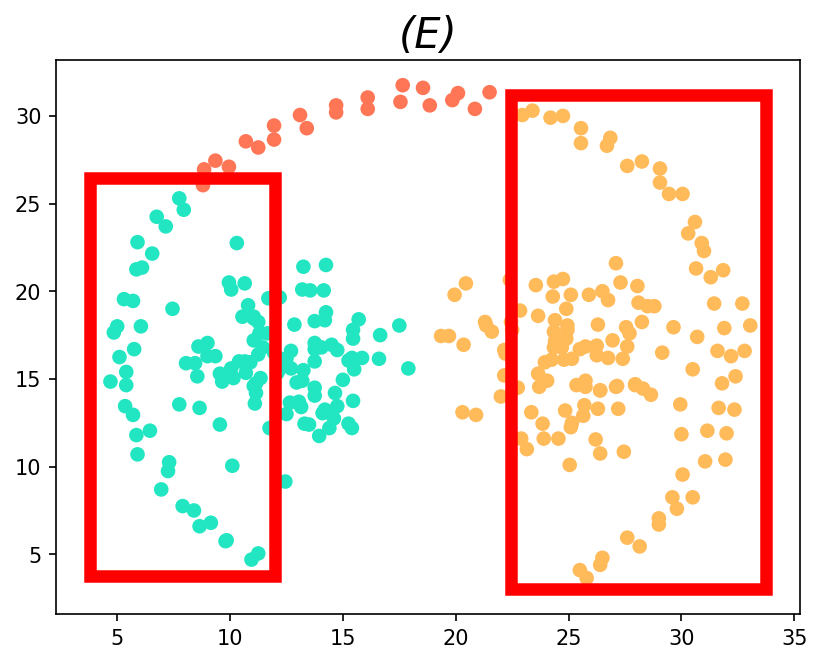}
  \includegraphics[width=1.1in]{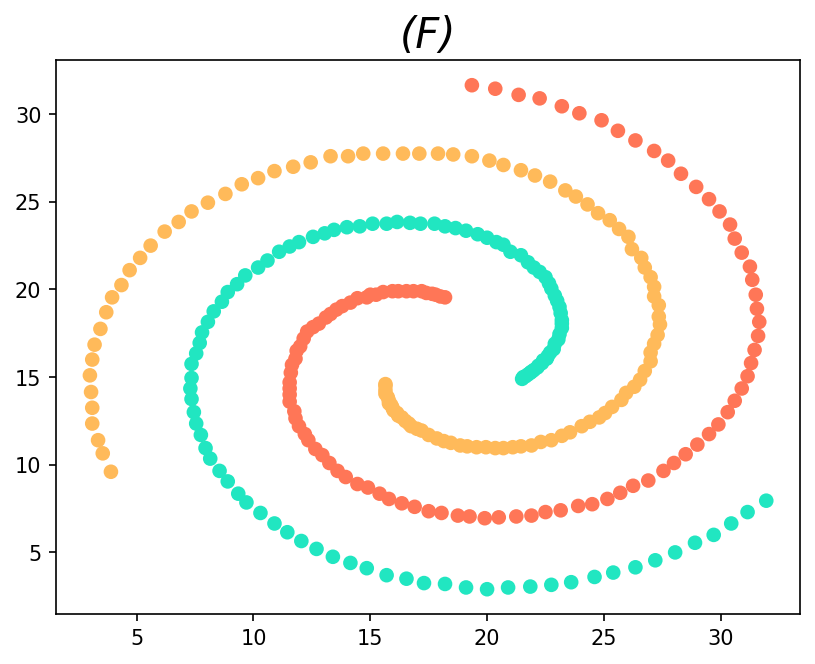}\\
  \includegraphics[width=1.1in]{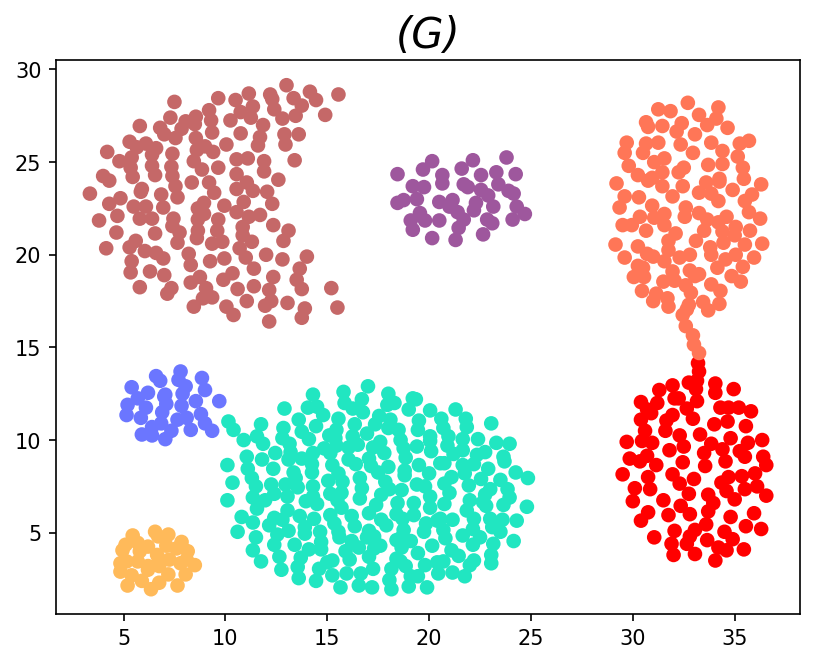}
  \includegraphics[width=1.1in]{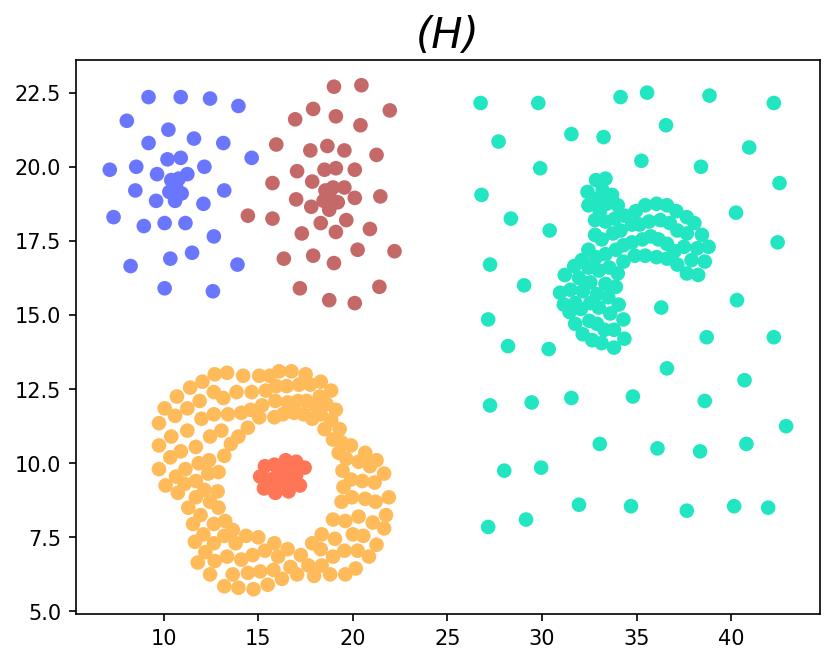}
  \includegraphics[width=1.1in]{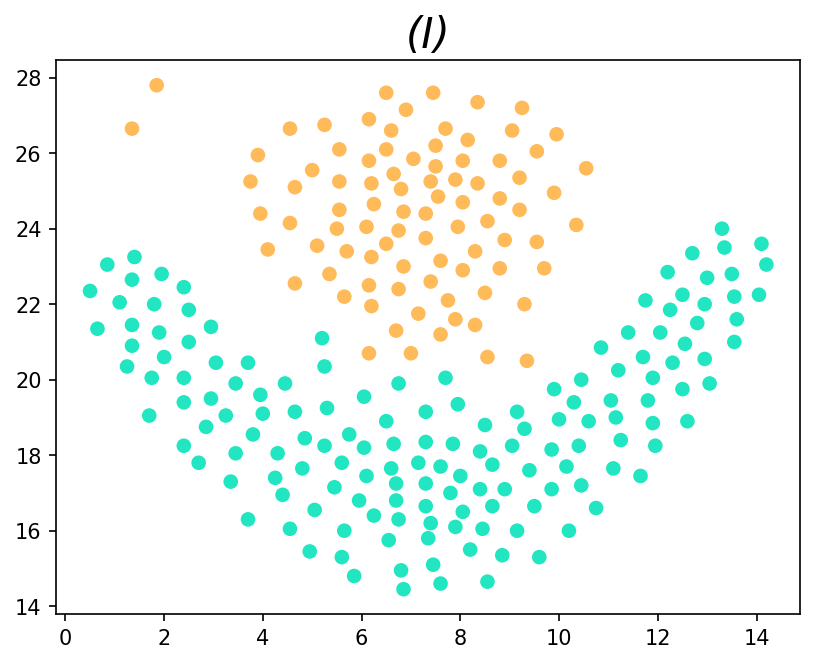}
  \includegraphics[width=1.1in]{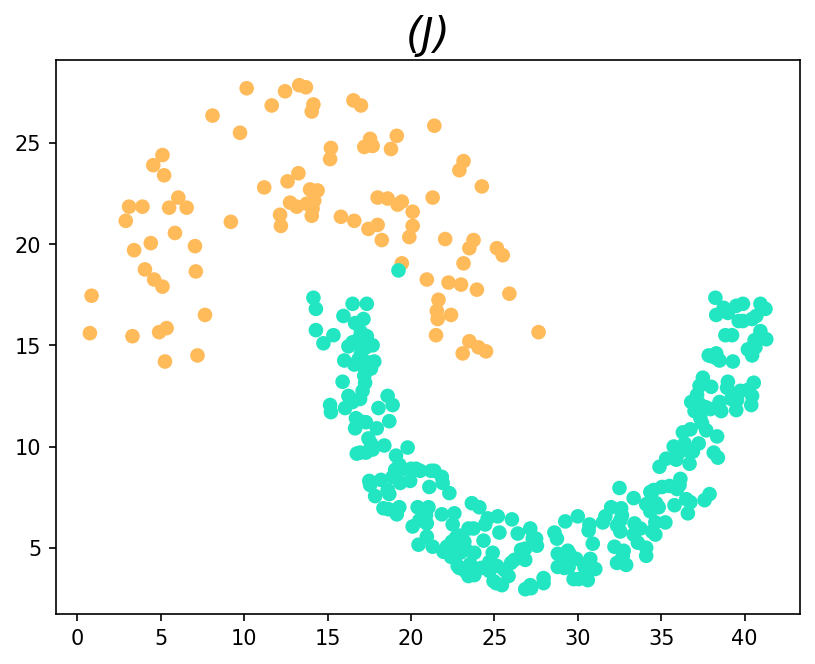}
  \includegraphics[width=1.1in]{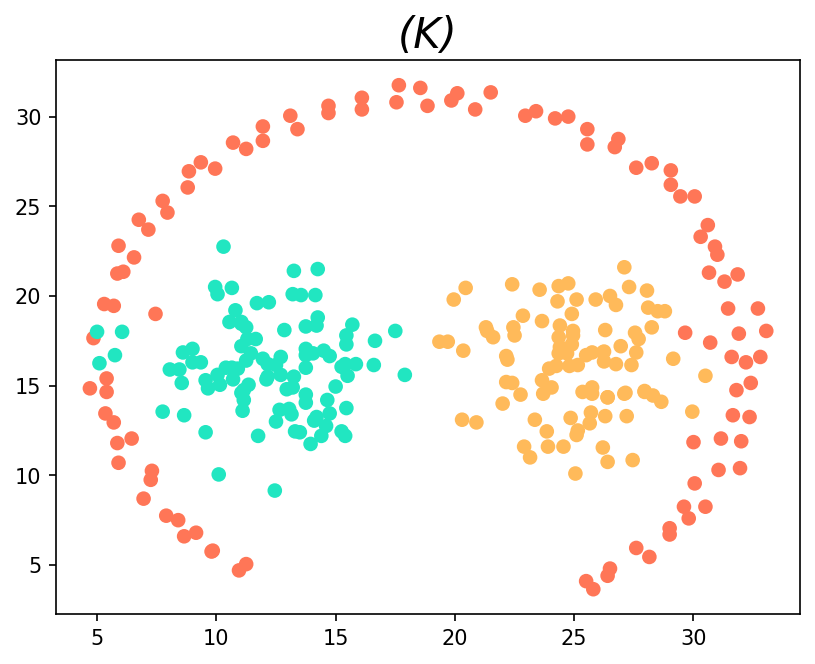}
  \includegraphics[width=1.1in]{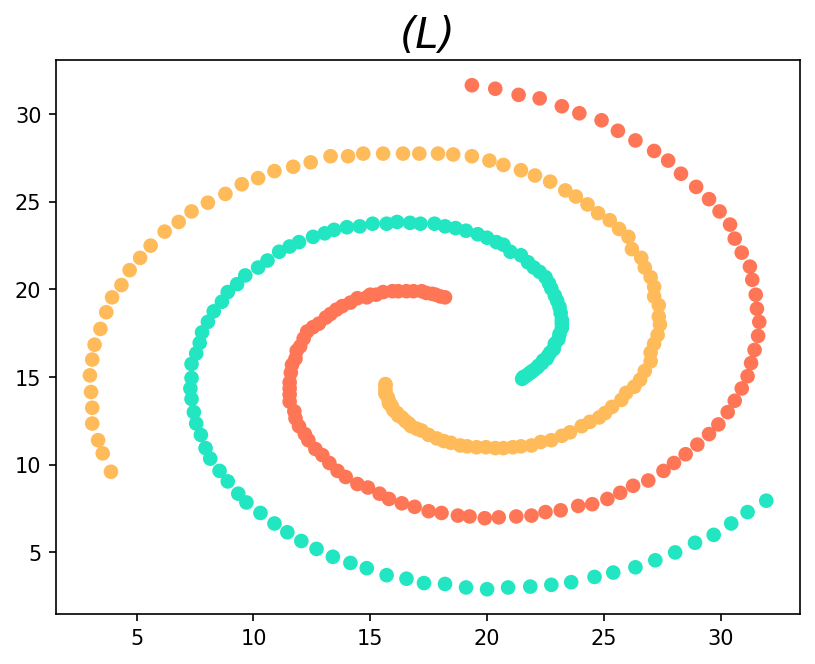}
  \caption{\textbf{The original clustering results (row 1) and the ECAC-optimized clustering results (row 2) of DPC on \emph{aggregation} (A, G), \emph{compound} (B, H), \emph{flame} (C, I), \emph{jain} (D, J), \emph{pathbased} (E, K), and \emph{spiral} (F, L):} The colors indicate the categories identified by DPC. DPC performs poorly on \emph{jain} and \emph{pathbased}. After ECAC optimization, DPC accurately identifies the vast majority of objects, in each cluster of \emph{jain} and \emph{pathbased}, as one category.}
  \label{fig:compare-dpc}
\end{figure*}

\begin{figure*}
  \centering
  \includegraphics[width=1.1in]{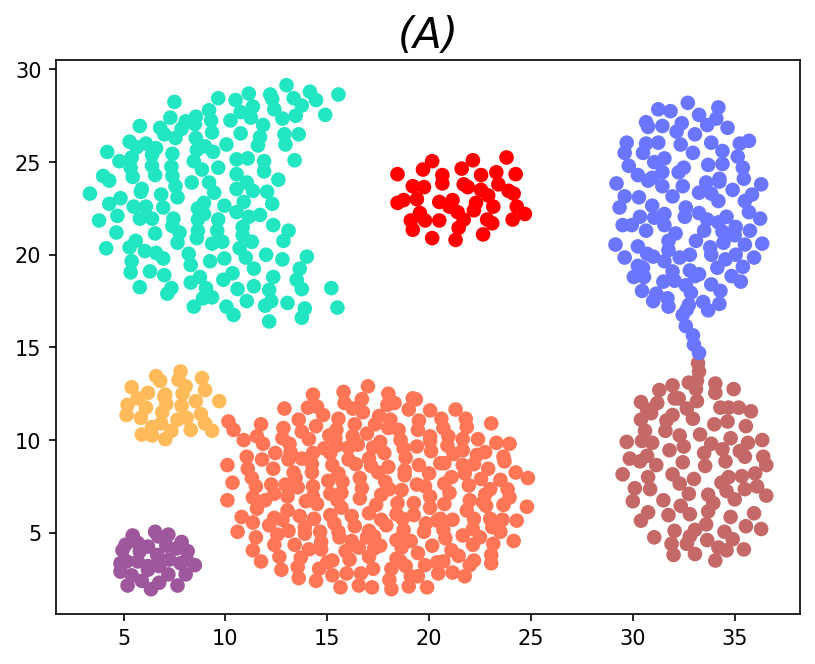}
  \includegraphics[width=1.1in]{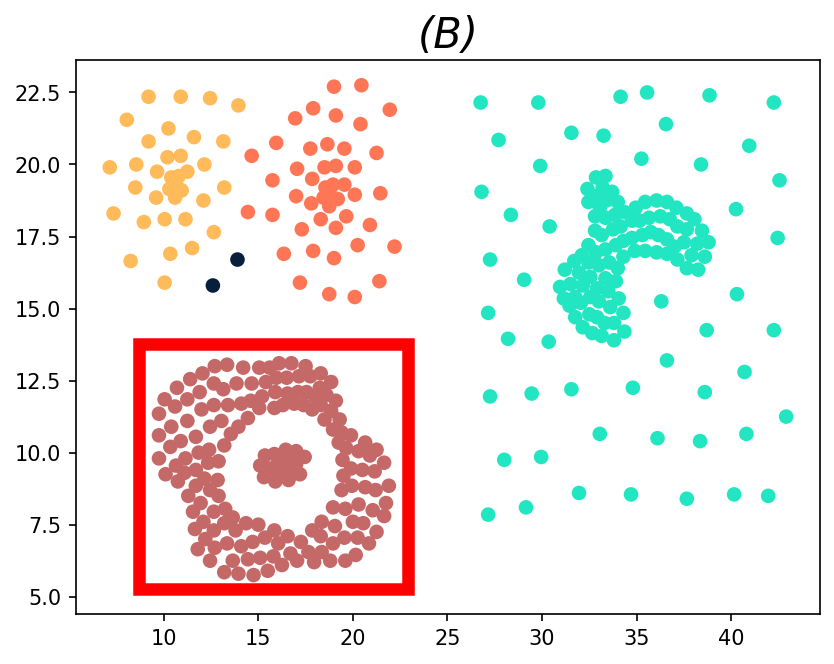}
  \includegraphics[width=1.1in]{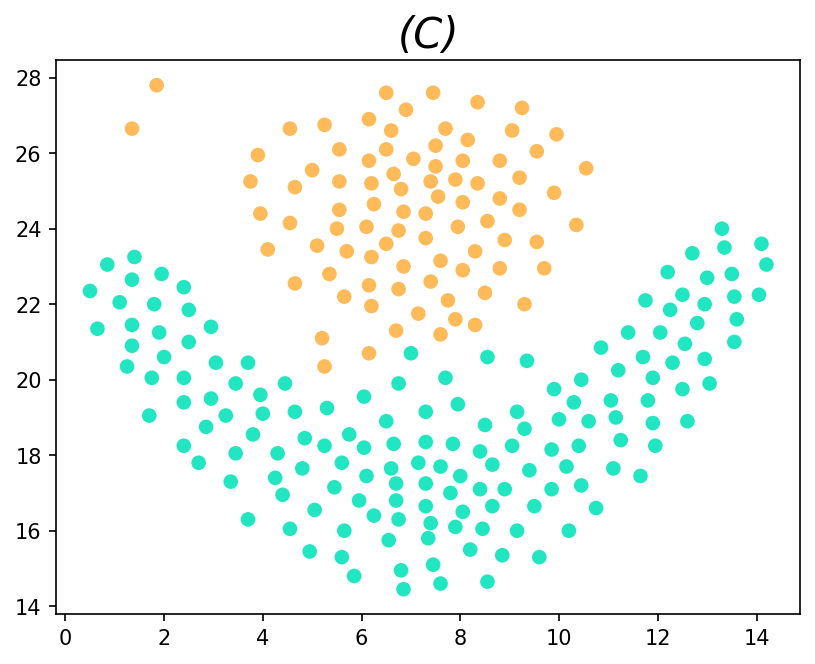}
  \includegraphics[width=1.1in]{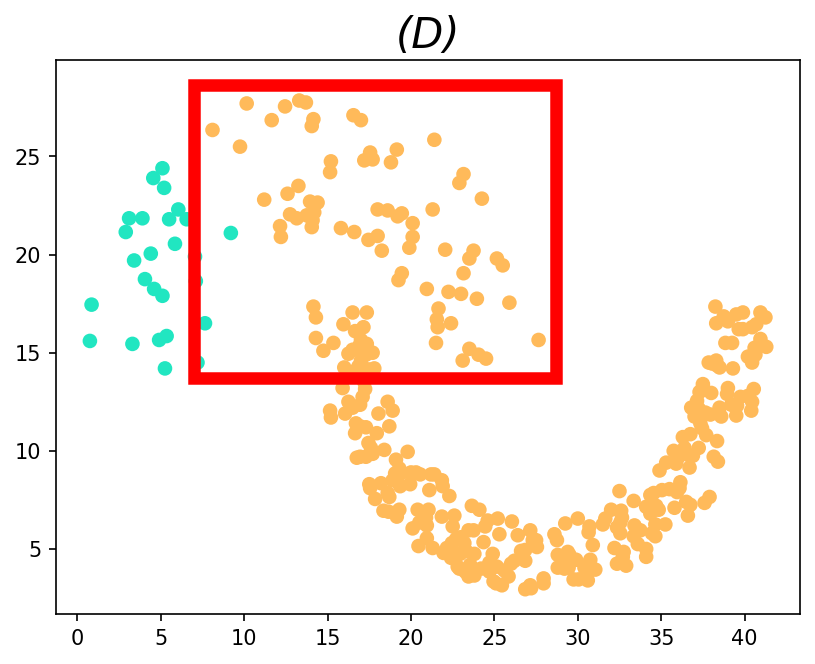}
  \includegraphics[width=1.1in]{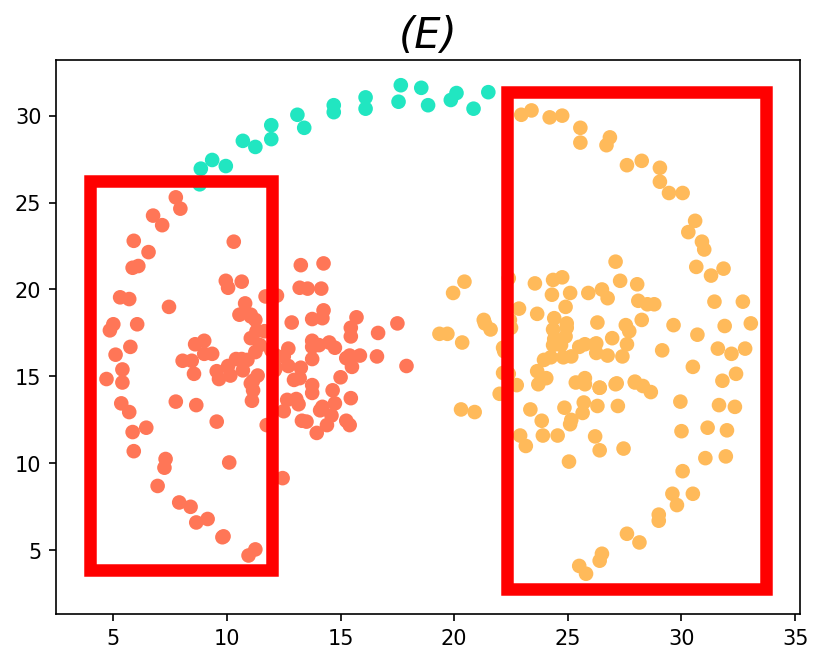}
  \includegraphics[width=1.1in]{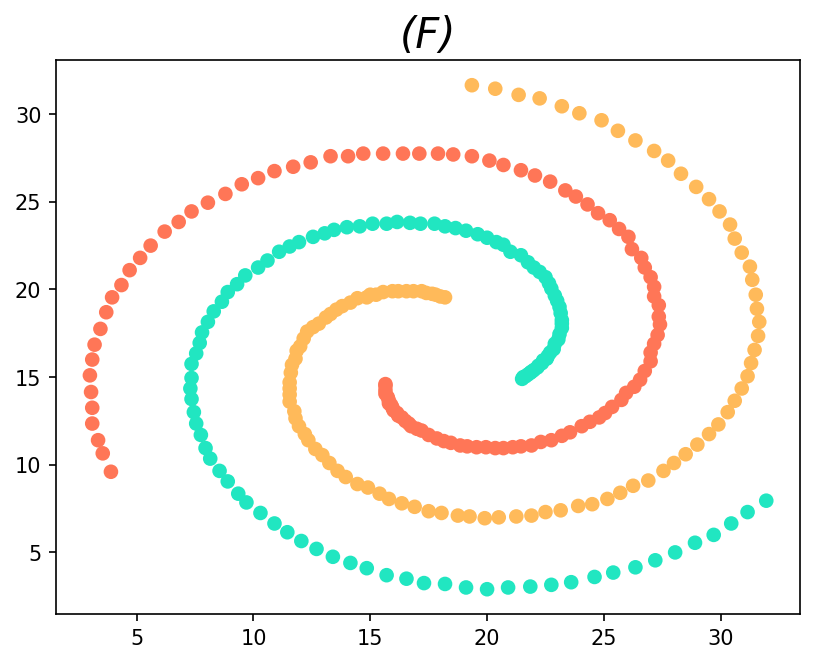}\\
  \includegraphics[width=1.1in]{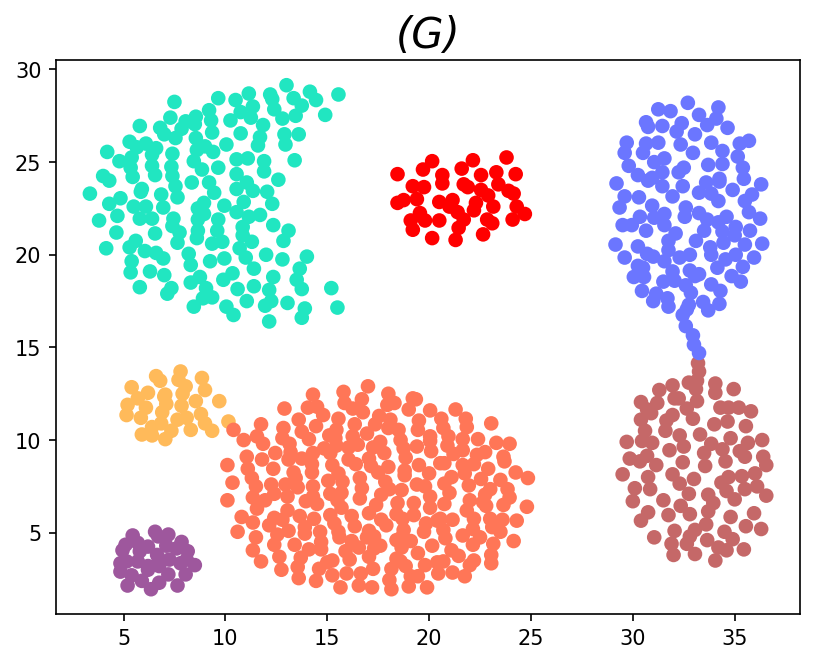}
  \includegraphics[width=1.1in]{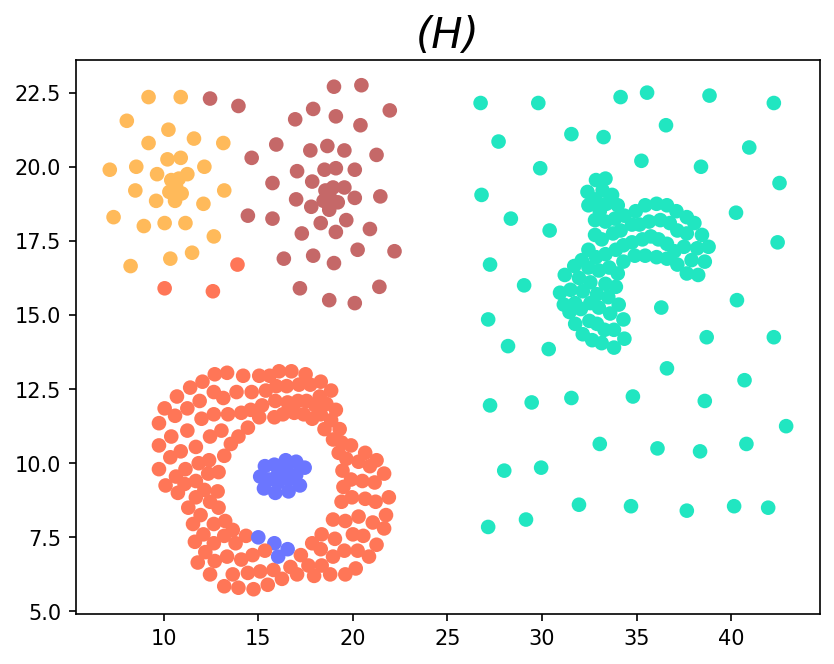}
  \includegraphics[width=1.1in]{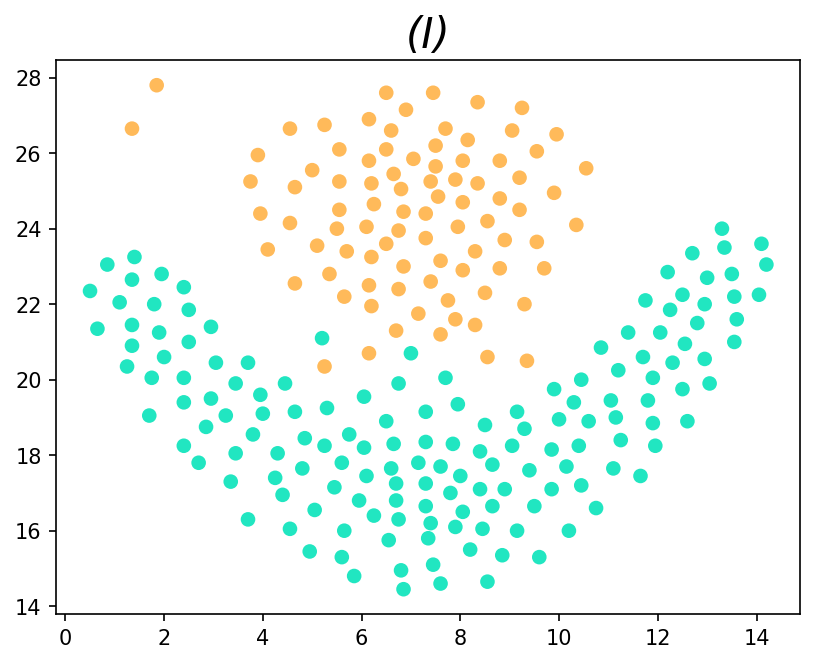}
  \includegraphics[width=1.1in]{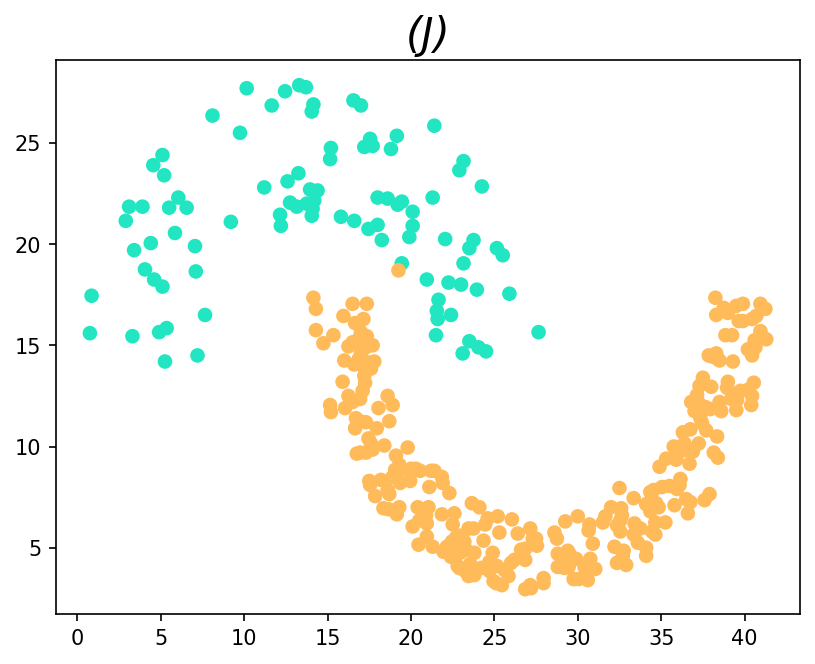}
  \includegraphics[width=1.1in]{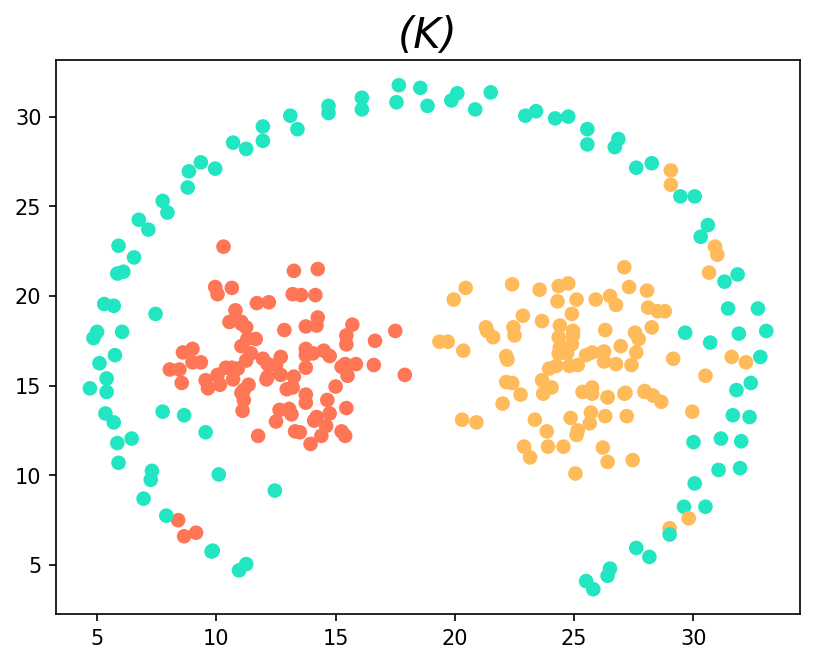}
  \includegraphics[width=1.1in]{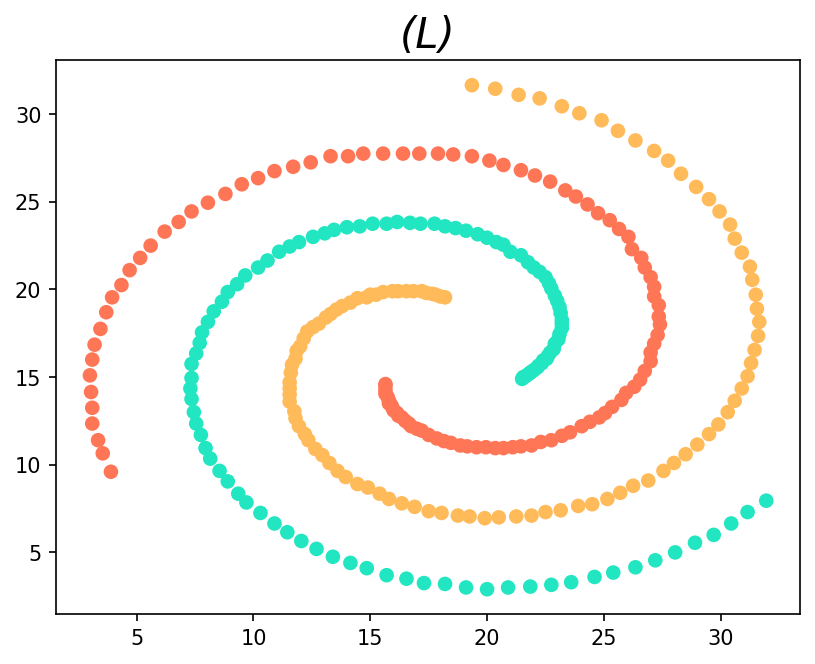}
  \caption{\textbf{The original clustering results (row 1) and the ECAC-optimized clustering results (row 2) of Extreme clustering on \emph{aggregation} (A, G), \emph{compound} (B, H), \emph{flame} (C, I), \emph{jain} (D, J), \emph{pathbased} (E, K), and \emph{spiral} (F, L):} The colors indicate the categories identified by Extreme clustering. Extreme clustering performs poorly on \emph{compound}, \emph{jain}, and \emph{pathbased}. After ECAC optimization, Extreme clustering accurately identifies the vast majority of objects, in each cluster of \emph{compound}, \emph{jain}, and \emph{pathbased}, as one category.}
  \label{fig:compare-extreme}
\end{figure*}

\subsubsection{Robustness to clustering centers}
\label{sec:generality}
In this paper, ECAC will be only used to optimize DPC, K-means, and Extreme clustering (see Section \ref{sec:Comparison} for details), but it does not mean that ECAC can only optimize them. Here, we will discuss whether ECAC is a general method that can optimize different center-based clustering algorithms. 

The difference between center-based clustering algorithms is mainly reflected in the difference between identified clustering centers. Therefore, as long as we confirm that the optimization effect of ECAC is independent of which objects serve as clustering centers, we can conclude that ECAC is a general method. We test ECAC on \emph{aggregation}, \emph{compound}, \emph{flame}, \emph{jain}, \emph{pathbased}, \emph{spiral}, \emph{skewed}, \emph{t4.8k}, \emph{t7.10k}, and \emph{t8.8k} datasets. For each dataset, we randomly select 15 disjoint groups of clustering centers, each of which is input to ECAC to obtain a set of extended-centers. Next, the extended-centers derived from different groups of clustering centers are separately input to the category assignment process of DPC to obtain different clustering results. Figure \ref{fig:centers} shows the clustering accuracy (NMI as well as RI) under different groups of clustering centers. The experimental results show that the accuracy is very stable regardless of which objects serve as clustering centers, indicating that the optimization effect of ECAC is not related to clustering centers. Therefore, ECAC is a general method and can optimize different center-based clustering algorithms.

\subsection{Comparison experiments}
\label{sec:Comparison}
In this subsection, we will compare the clustering accuracy of K-means, DPC, and Extreme clustering before and after ECAC optimization.

\subsubsection{Comparison on synthetic datasets}
We select some synthetic datasets that are difficult to be clustered, \emph{i.e.}, \emph{aggregation}, \emph{compound}, \emph{flame}, \emph{jain}, \emph{pathbased}, and \emph{spiral}. Figure \ref{fig:compare-dpc} and Figure \ref{fig:compare-extreme} show the original clustering results (row 1) and the ECAC-optimized clustering results (row 2) of DPC and Extreme clustering on these synthetic datasets, where the colors indicate the identified categories. The experimental results show that DPC performs poorly on \emph{jain} and \emph{pathbased}, and Extreme clustering performs poorly on \emph{compound}, \emph{jain}, and \emph{pathbased} (see red boxes for details). Specifically, they cannot correctly identify the upper cluster of \emph{jain} and the cyclic cluster of \emph{pathbased}. Many objects in these 2 clusters are incorrectly marked in the same color as adjacent clusters. Extreme clustering also fails to distinguish between the nested clusters of \emph{compound} (\emph{i.e.}, the cyclic cluster and the Gaussian cluster). After ECAC optimization, DPC and Extreme clustering successfully identify them. Only a few objects are misidentified due to the category assignment strategy of DPC and Extreme clustering.

\begin{figure*}
  \centering
  \includegraphics[width=1.1in]{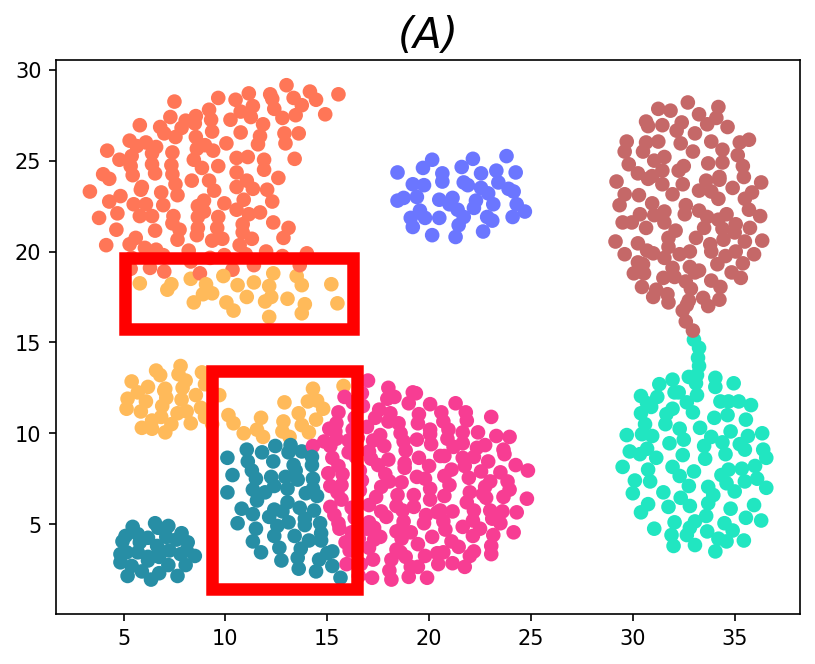}
  \includegraphics[width=1.1in]{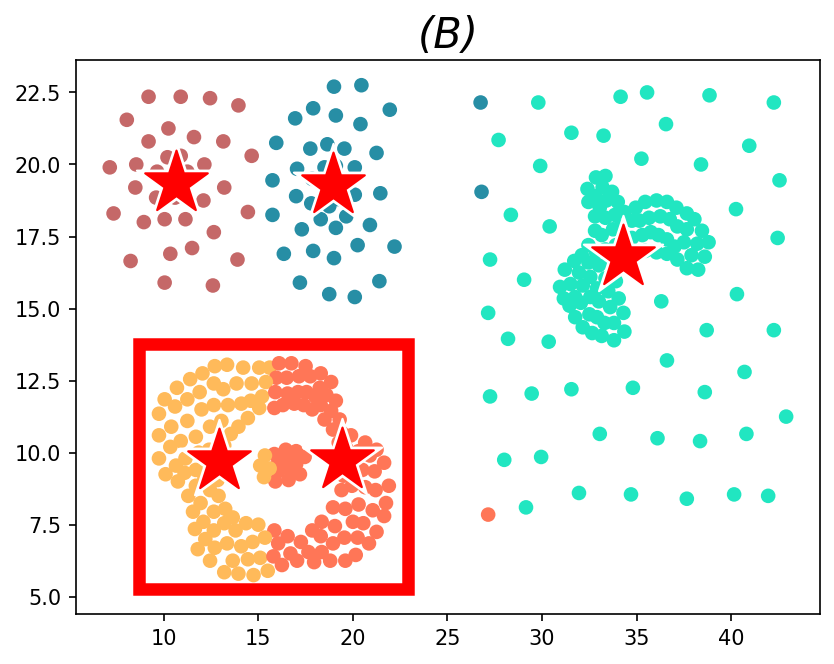}
  \includegraphics[width=1.1in]{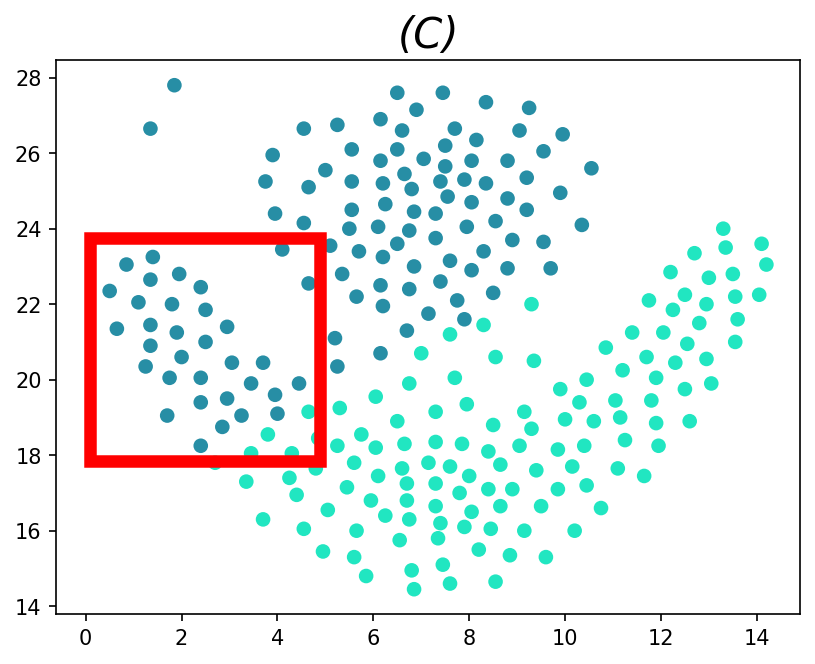}
  \includegraphics[width=1.1in]{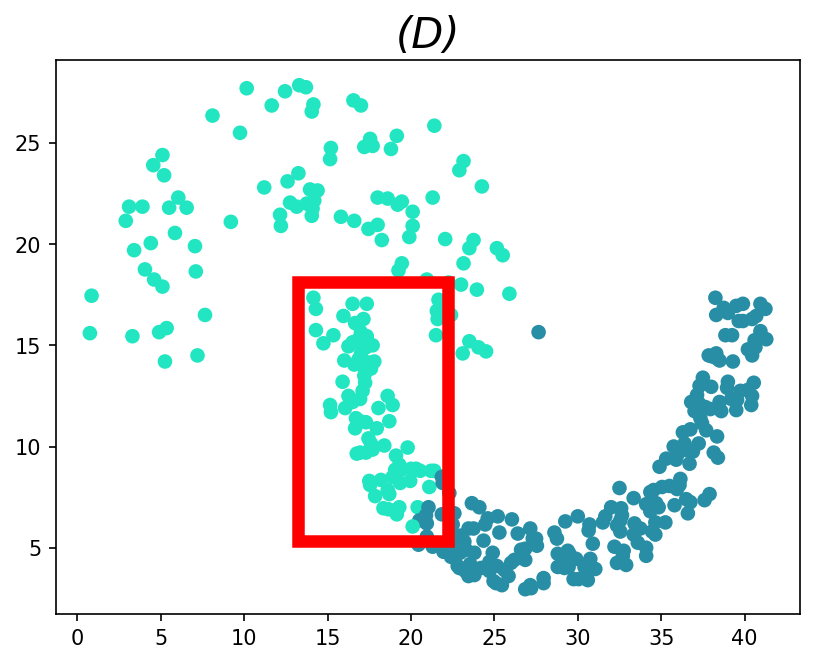}
  \includegraphics[width=1.1in]{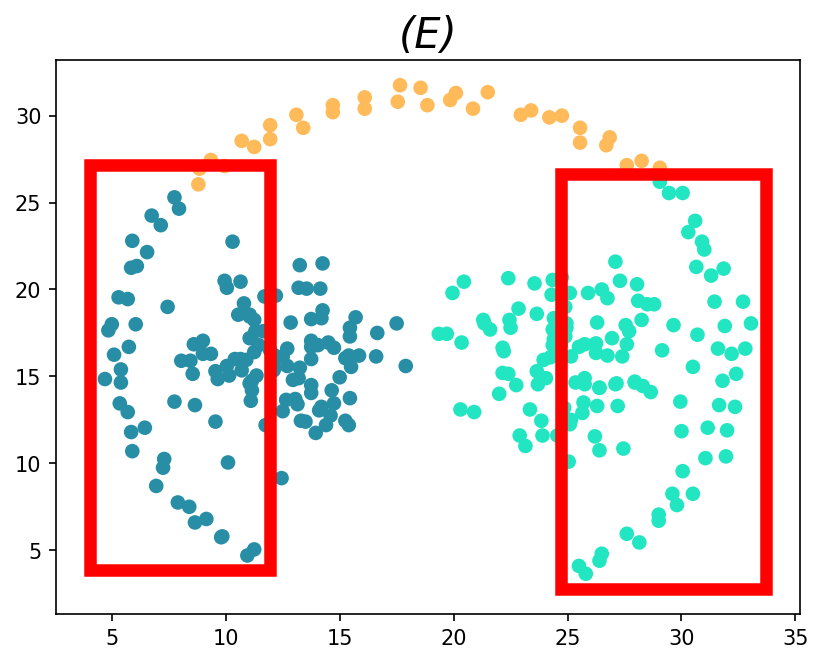}
  \includegraphics[width=1.1in]{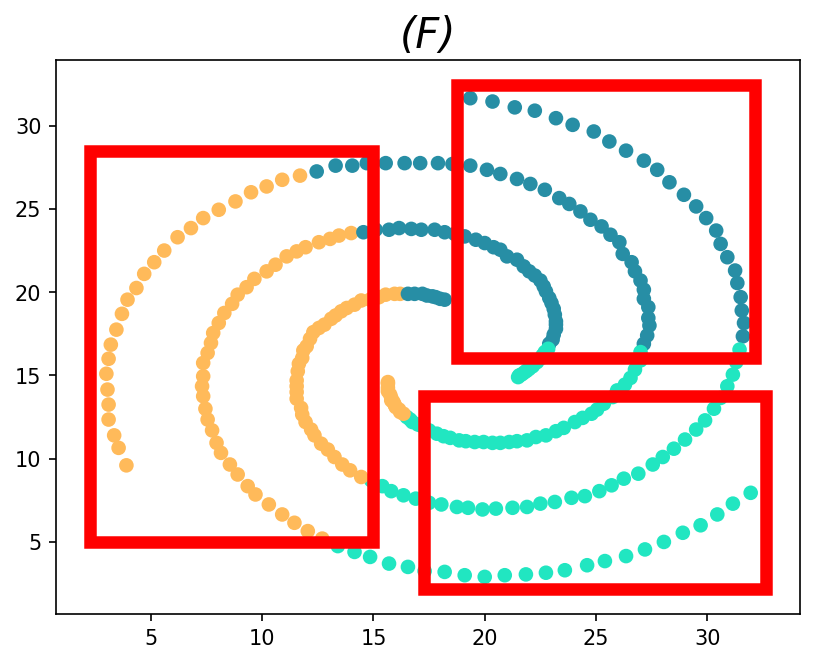}\\
  \includegraphics[width=1.1in]{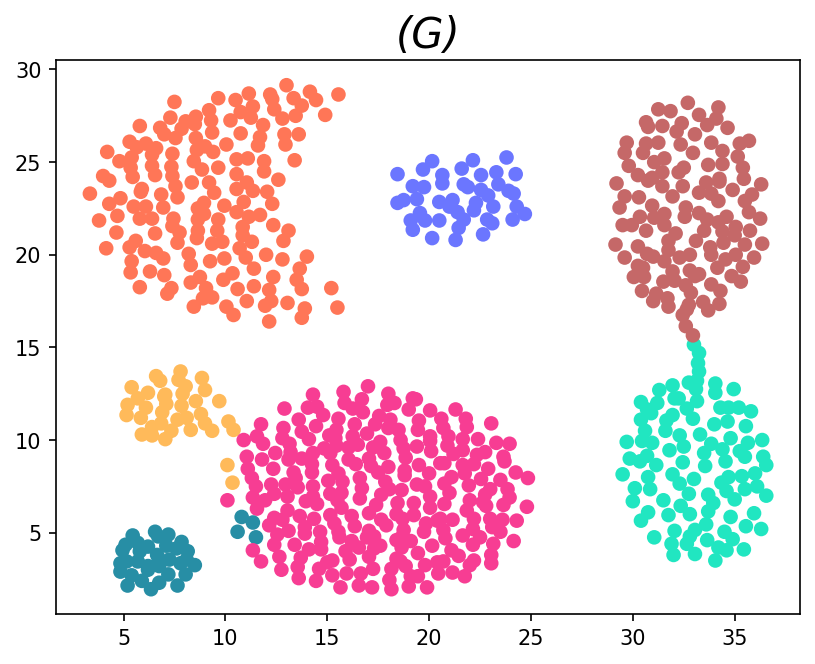}
  \includegraphics[width=1.1in]{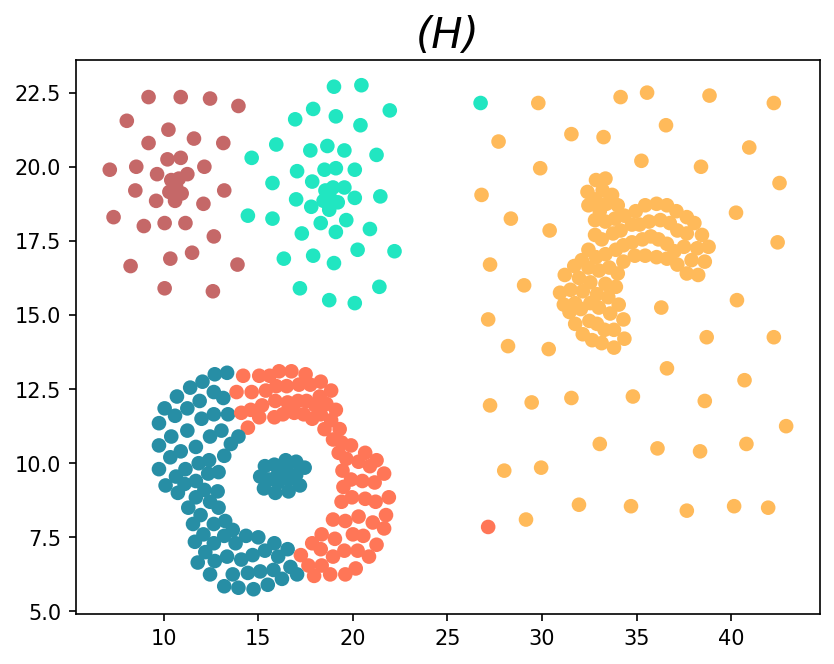}
  \includegraphics[width=1.1in]{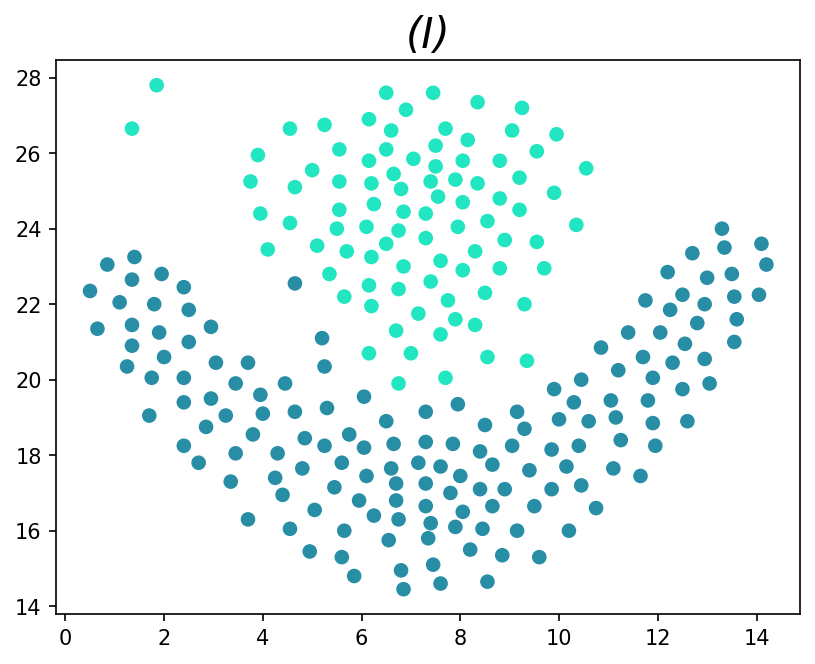}
  \includegraphics[width=1.1in]{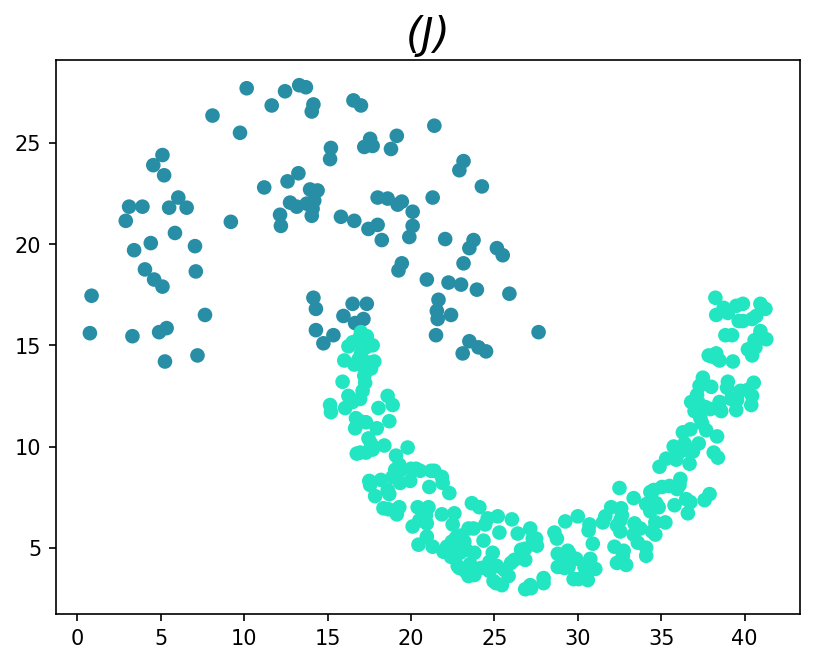}
  \includegraphics[width=1.1in]{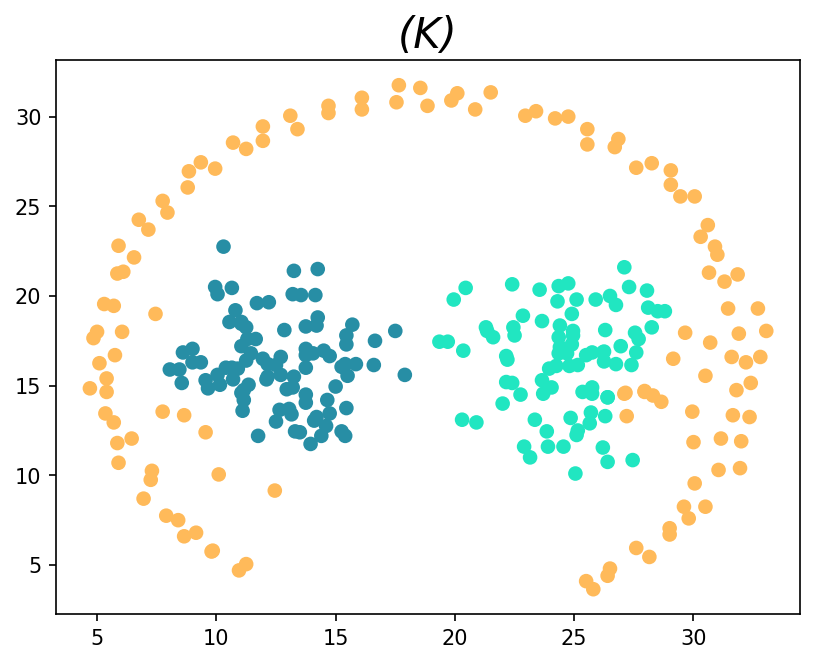}
  \includegraphics[width=1.1in]{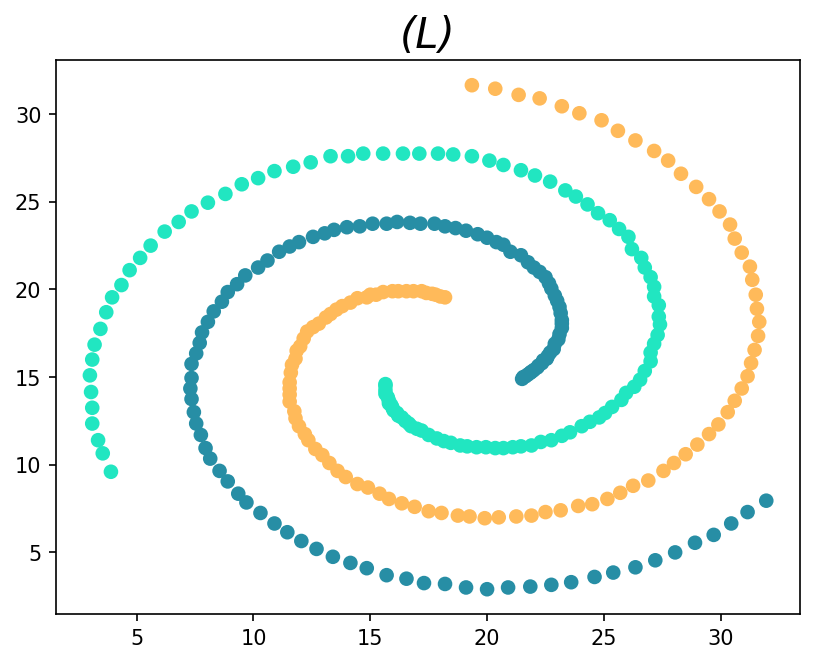}
  \caption{\textbf{The original clustering results (row 1) and the ECAC-optimized clustering results (row 2) of K-means on \emph{aggregation} (A, G), \emph{compound} (B, H), \emph{flame} (C, I), \emph{jain} (D, J), \emph{pathbased} (E, K), and \emph{spiral} (F, L):} The colors indicate the categories identified by K-means. K-means performs poorly on all datasets. After ECAC optimization, K-means accurately identifies shaped clusters, and its clustering accuracy is greatly improved.}
  \label{fig:compare-kmeans}
\end{figure*}

Figure \ref{fig:compare-kmeans} shows the original clustering results (row 1) and the ECAC-optimized clustering results (row 2) of K-means, where the colors indicate the categories identified by K-means. The experimental results show that K-means performs poorly on all datasets (see red boxes for details). After ECAC optimization, except on \emph{compound}, the accuracy of K-means on other datasets is improved significantly. The reason why ECAC cannot optimize the result of K-means on \emph{compound} is that K-means incorrectly identifies 2 clustering centers in the cyclic cluster but no clustering centers in the Gaussian cluster (see the pentagons in Figure \ref{fig:compare-kmeans}(B) for details), resulting in the extended-sets not corresponding to the clusters one by one.

\begin{table*}
\caption{Accuracy of center-based clustering algorithms before and after ECAC optimization.}
\label{tab:Accuracy}
    \setlength\tabcolsep{3pt}
    \centering
    \begin{tabular}{c|c|cccccccccc}
        \hline
        \multicolumn{2}{c}{} &
        \emph{dermatology}
                             &
        \emph{movement}
                             &
        \emph{optdigits}
                             &
        \emph{page}
                             &
        \emph{penbased}
                             &
        \emph{texture}
                             &
        \emph{satimage}
                             &
        \emph{breast}
                             &
        \emph{wifi\_loc}
                             &
        \emph{banknote}
        \\

        \hline

        \multirow{9}{*}{NMI} &
        K-means
                             &
        0.090
                             &
        0.54
                             &
        0.75
                             &
        0.052
                             &
        0.69
                             &
        0.63
                             &
        0.63
                             &
        0.46
                             &
        0.89
                             &
        0.030
        \\

                             & DPC
                             &
        0.23
                             &
        0.54
                             &
        0.75
                             &
        0.0022
                             &
        0.74
                             &
        0.71
                             &
        0.57
                             &
        0.31
                             &
        0.86
                             &
        0.35
        \\

                             & Extreme
                             &
        0.41
                             &
        0.47
                             &
        0.83
                             &
        0.083
                             &
        0.73
                             &
        0.70
                             &
        0.42
                             &
        0.0017
                             &
        0.83
                             &
        0.51
        \\
        \cline{2-12}

                             & K-means+ECAC
                             &
        \makecell{0.31                      \\
            \textcolor{red}{ (+242.3\%)}}
                             &
        \makecell{0.60                      \\
            \textcolor{red}{ (+9.9\%)}}
                             &
        \makecell{0.86                      \\
            \textcolor{red}{ (+13.7\%)}}
                             &
        \makecell{0.058                     \\
            \textcolor{red}{ (+11.6\%)}}
                             &
        \makecell{0.76                      \\
            \textcolor{red}{ (+10.3\%)}}
                             &
        \makecell{0.76                      \\
            \textcolor{red}{ (+19.7\%)}}
                             &
        \makecell{0.64                      \\
            \textcolor{red}{ (+1.5\%)}}
                             &
        \makecell{0.52                      \\
            \textcolor{red}{ (+12.6\%)}}
                             &
        \makecell{0.92                      \\
            \textcolor{red}{ (+3.3\%)}}
                             &
        \makecell{0.12                      \\
            \textcolor{red}{ (+310.6\%)}}
        \\

                             & DPC+ECAC
                             &
        \makecell {  0.36                   \\
            \textcolor{red}{ (+57.3\%)}
        }                    &
        \makecell {  0.51                   \\
            \textcolor{blue}{($-$6.5\%)}
        }                    &
        \makecell {   0.80                  \\
            \textcolor{red}{ (+6.6\%)}
        }                    &
        \makecell { 0.049                   \\
            \textcolor{red}{ (+2088.0\%)}
        }                    &
        \makecell {   0.74                  \\
            \textcolor{red}{ (+0.1\%)}
        }                    &
        \makecell {   0.76                  \\
            \textcolor{red}{ (+6.7\%)}
        }                    &
        \makecell {   0.60                  \\
            \textcolor{red}{ (+6.9\%)}
        }                    &
        \makecell {  0.38                   \\
            \textcolor{red}{ (+21.0\%)}
        }                    &
        \makecell {   0.91                  \\
            \textcolor{red}{ (+6.0\%)}
        }                    &
        \makecell {  0.56                   \\
            \textcolor{red}{ (+60.0\%)}
        }                                   \\

                             & Extreme+ECAC
                             &
        \makecell {   0.43                  \\
            \textcolor{red}{ (+4.5\%)}
        }                    &
        \makecell {  0.53                   \\
            \textcolor{red}{ (+12.6\%)}
        }                    &
        \makecell {   0.84                  \\
            \textcolor{red}{ (+1.4\%)}
        }                    &
        \makecell {  0.18                   \\
            \textcolor{red}{ (+110.5\%)}
        }                    &
        \makecell {   0.79                  \\
            \textcolor{red}{ (+8.2\%)}
        }                    &
        \makecell {   0.76                  \\
            \textcolor{red}{ (+7.7\%)}
        }                    &
        \makecell {  0.49                   \\
            \textcolor{red}{ (+16.8\%)}
        }                    &
        \makecell {  0.38                   \\
            \textcolor{red}{ (+22149.5\%)}
        }                    &
        \makecell {   0.85                  \\
            \textcolor{red}{ (+2.5\%)}
        }                    &
        \makecell {  0.50                   \\
            \textcolor{blue}{($-$1.1\%)}
        }
        \\ \hline
        \hline

        \multirow{9}{*}{RI}  &
        K-means
                             &
        0.028
                             &
        0.31
                             &
        0.67
                             &
        0.013
                             &
        0.60
                             &
        0.46
                             &
        0.57
                             &
        0.49
                             &
        0.89
                             &
        0.049
        \\

                             & DPC
                             &
        0.10
                             &
        0.29
                             &
        0.56
                             &
        0.0020
                             &
        0.55
                             &
        0.48
                             &
        0.47
                             &
        0.28
                             &
        0.85
                             &
        0.23
        \\

                             & Extreme
                             &
        0.17
                             &
        0.23
                             &
        0.69
                             &
        0.0047
                             &
        0.52
                             &
        0.43
                             &
        0.30
                             &
        0.0024
                             &
        0.82
                             &
        0.39
        \\ \cline{2-12}

                             & K-means+ECAC
                             &
        \makecell {  0.25                   \\
            \textcolor{red}{ (+803.4\%)}
        }                    &
        \makecell {  0.37                   \\
            \textcolor{red}{ (+17.1\%)}
        }                    &
        \makecell {  0.80                   \\
            \textcolor{red}{ (+19.6\%)}
        }                    &
        \makecell { 0.047                   \\
            \textcolor{red}{ (+260.9\%)}
        }                    &
        \makecell {   0.65                  \\
            \textcolor{red}{ (+9.8\%)}
        }                    &
        \makecell {  0.56                   \\
            \textcolor{red}{ (+21.1\%)}
        }                    &
        \makecell {   0.60                  \\
            \textcolor{red}{ (+4.8\%)}
        }                    &
        \makecell {  0.60                   \\
            \textcolor{red}{ (+22.2\%)}
        }                    &
        \makecell {   0.94                  \\
            \textcolor{red}{ (+5.7\%)}
        }                    &
        \makecell {  0.10                   \\
            \textcolor{red}{ (+115.9\%)}
        }                                   \\

                             & DPC+ECAC
                             &
        \makecell {  0.29                   \\
            \textcolor{red}{ (+190.9\%)}
        }                    &
        \makecell {  0.26                   \\
            \textcolor{blue}{($-$11.5\%)}
        }                    &
        \makecell {  0.67                   \\
            \textcolor{red}{ (+19.3\%)}
        }                    &
        \makecell { 0.065                   \\
            \textcolor{red}{ (+3093.4\%)}
        }                    &
        \makecell {   0.59                  \\
            \textcolor{red}{ (+7.8\%)}
        }                    &
        \makecell {  0.55                   \\
            \textcolor{red}{ (+14.2\%)}
        }                    &
        \makecell {  0.52                   \\
            \textcolor{red}{ (+12.6\%)}
        }                    &
        \makecell {  0.40                   \\
            \textcolor{red}{ (+40.3\%)}
        }                    &
        \makecell {  0.93                   \\
            \textcolor{red}{ (+10.2\%)}
        }                    &
        \makecell {  0.55                   \\
            \textcolor{red}{ (+136.0\%)}
        }                                   \\

                             & Extreme+ECAC
                             &
        \makecell {   0.18                  \\
            \textcolor{red}{ (+7.1\%)}
        }                    &
        \makecell {  0.31                   \\
            \textcolor{red}{ (+34.5\%)}
        }                    &
        \makecell {  0.82                   \\
            \textcolor{red}{ (+18.0\%)}
        }                    &
        \makecell {  0.15                   \\
            \textcolor{red}{ (+3096.4\%)}
        }                    &
        \makecell {  0.71                   \\
            \textcolor{red}{ (+36.9\%)}
        }                    &
        \makecell {  0.53                   \\
            \textcolor{red}{ (+24.6\%)}
        }                    &
        \makecell {  0.40                   \\
            \textcolor{red}{ (+35.7\%)}
        }                    &
        \makecell {  0.37                   \\
            \textcolor{red}{ (+15203.0\%)}
        }                    &
        \makecell {   0.87                  \\
            \textcolor{red}{ (+7.2\%)}
        }                    &
        \makecell {  0.38                   \\
            \textcolor{blue}{($-$3.0\%)}
        }
        \\ \hline
    \end{tabular}
\end{table*}

\subsubsection{Comparison on real-world datasets}
We compute the original accuracy and the ECAC-optimized accuracy of K-means, DPC, and Extreme clustering on 10 real-world datasets, \emph{i.e.}, \emph{dermatology}, \emph{movement}, \emph{optdigits}, \emph{page}, \emph{penbased}, \emph{texture}, \emph{satimage}, \emph{breast}, \emph{wifi\_loc}, and \emph{banknote}. The accuracy is recorded in Table \ref{tab:Accuracy}. Table \ref{tab:Accuracy} consists of 2 parts, with NMI as the accuracy measure in part 1 and RI as the accuracy measure in part 2. In addition, we also compute the improvement rates (\emph{i.e.}, (the ECAC-optimized accuracy - the original accuracy)/the original accuracy), recorded in the parentheses in Table \ref{tab:Accuracy}. If the accuracy is improved, the improvement rate is marked in red; otherwise, the improvement rate is marked in blue. The experimental results show that ECAC successfully improves the majority of clustering accuracy, and most of the improvement rates exceed 15\%. It only reduces the accuracy of DPC on \emph{movement} and Extreme clustering on \emph{banknote}, but the decrease rate is not significant. Specifically, the original accuracy (NMI) of K-means on \emph{dermatology} is only 0.09. After ECAC optimization, this accuracy is improved to 0.31. Although \emph{wifi\_loc} is an extremely clustable dataset, no clustering algorithm has a clustering accuracy of more than 0.9 on it. After ECAC optimization, the accuracy (NMI as well as RI) of K-means and DPC exceed 0.9. On \emph{page}, one of the most difficult datasets to be clustered, the original accuracy (RI) of Extreme clustering is close to 0. After ECAC optimization, this accuracy is improved to 0.15. The original accuracy (NMI) of Extreme clustering on \emph{breast} is also close to 0. After ECAC optimization, this accuracy is improved to 0.38. Excluding the clustering results on the two datasets with unusually high improvement rates (\emph{i.e.}, \emph{page} and \emph{breast}), ECAC improves the clustering accuracy (NMI) by 33.4\% on average and the clustering accuracy (RI) by 64.1\% on average. In summary, we can conclude that ECAC is an effective optimization method to optimize center-based clustering algorithms.

\section{Conclusion}
\label{sec:Conclusion}
Traditional center-based clustering algorithms are frequently used in real-world scenarios. However, they are difficult to achieve excellent clustering results on the datasets with the complex distribution. We believe the main reason is that a single clustering center in each cluster lacks sufficient representative capability to strongly represent distant objects. In this paper, we propose a novel optimization method called ECAC. It identifies some specific high-density objects as the extended-centers of the clustering centers until the extended-centers are uniformly spread throughout the clusters. These extended-centers will act as relays to expand the representative capability of the clustering centers, thus improving the accuracy of center-based clustering algorithms. More importantly, ECAC is independent of the center process and the category assignment process of center-based clustering algorithms, so it is a general optimization method that can optimize different algorithms. We conducted a series of ablation experiments to verify the necessity of the components of ECAC. We also conducted numerous experiments to test the robustness of ECAC. The experimental results show that ECAC is robust to diverse datasets. No matter how complex the dataset is, each clustering center is within the same cluster as its derived extended-centers, ensuring that the objects within the same cluster can be identified as one category. Moreover, ECAC is robust to diverse clustering centers. Its optimization effect remains stable regardless of which objects are selected as clustering centers. Finally, we compared the accuracy of center-based clustering algorithms on synthetic and real-world datasets before and after ECAC optimization. After ECAC optimization, the center-based clustering algorithms, which always fail to identify shaped clusters, accurately identify complex-shaped clusters. Compared with original clustering accuracy, the ECAC-optimized accuracy (NMI as well as RI) improves by an average of 33.4\% and 64.1\%, respectively.

\ifCLASSOPTIONcaptionsoff
  \newpage
\fi

\bibliographystyle{IEEEtran}

\bibliography{ECAC}

\end{document}